\pdfoutput=1
\documentclass[9pt,shortpaper,twoside,web]{ieeecolor}

\usepackage{etoolbox}
\makeatletter
\@ifundefined{color@begingroup}%
{\let\color@begingroup\relax\let\color@endgroup\relax}{}%
\def\fix@ieeecolor@hbox#1{%
\hbox{\color@begingroup#1\color@endgroup}}
\patchcmd\@makecaption{\hbox}{\fix@ieeecolor@hbox}{}{\FAILED}
\patchcmd\@makecaption{\hbox}{\fix@ieeecolor@hbox}{}{\FAILED}

\usepackage{generic}
\usepackage{cite}
\usepackage{amsmath,amssymb,amsfonts}
\usepackage{algorithmic}
\usepackage{algorithm}
\usepackage{graphicx}
\usepackage{textcomp}
\usepackage{booktabs}
\usepackage{makecell}
\usepackage{multirow}
\usepackage{tabularx}
\usepackage{hyperref}

\makeatletter
\let\ps@titlepagestyle\ps@empty
\makeatother

\newcommand{\figref}[1]{Fig.~\ref{#1}}
\newcommand{\subsubsubsection}[1]{\textbf{#1:}}

\begin{document}
\title{Efficient 3D Whole-Body PET Image Denoising via Conditional Rectified Flow With Optimized Sampling Strategy}
\author{Jiale Shen, Guolin Wang, Chenhao Wang, Xinhui Su*, Wei Luo* and \IEEEmembership{Feng Yu}*
\thanks{Jiale Shen and Guolin Wang contributed equally to this work.}
\thanks{Corresponding authors: Xinhui Su; Wei Luo; Feng Yu.}
\thanks{Jiale Shen, Chenhao Wang, Wei Luo and Feng Yu are with the College of Biomedical Engineering \& Instrument Science, Zhejiang University, Zhejiang, 310063, China. (E-mails: \{jimmy.s, gordon\_won, luo.wei and osfengyu\}@zju.edu.cn)}
\thanks{Guolin Wang and Xinhui Su are with the Department of Nuclear Medicine, The First Affiliated Hospital, Zhejiang University School of Medicine, Zhejiang, 310003, China. (E-mails: \{wangguolin, suxinhui\}@zju.edu.cn)}
}
\maketitle
\begin{abstract}
Reducing radiation exposure in Positron Emission Tomography (PET) is important for patient safety; however, ultra-low-dose imaging suffers from severe noise, which may affect diagnostic interpretation without appropriate image enhancement.
While current 3D deep generative models, particularly diffusion models, have shown strong reconstruction fidelity, their practical use can be limited by long inference times.
In contrast, faster 2D-based alternatives may have difficulty maintaining volumetric consistency, an important consideration for whole-body PET imaging analysis.
To bridge this gap, we propose a one-pass conditional 3D rectified flow (3D Flow) framework for whole-body PET image denoising that incorporates a novel optimized non-uniform sampling strategy.
The model is trained with a one-pass linear-interpolant velocity-matching objective.
This approach reconstructs a full 3D volume in approximately 30 seconds in our implementation, compared with multi-hour inference for the evaluated 3D DDPM baseline.
Evaluations including zero-shot transfer to an independent clinical dataset show that our model achieves favorable global image quality and lesion conspicuity compared with the evaluated 3D DDPM and DDIM baselines, including on challenging short-acquisition data. 
Furthermore, the proposed method shows promising zero-shot transfer performance across the evaluated datasets and unseen dose levels (down to 1/100 of the standard dose), with artifact-focused visual comparisons supporting the need for further lesion-level validation.
By balancing reconstruction fidelity and computational efficiency, this work presents a candidate approach for ultra-low-dose whole-body PET image denoising.
\textit{Code is publicly available at: \url{https://anonymous.4open.science/r/PET-Rectified-Flow/}}
\end{abstract}


\section{Introduction}
\label{sec:introduction}
\IEEEPARstart{P}{ositron} emission tomography (PET) is a crucial quantitative functional imaging modality widely utilized across various clinical domains, including oncology, neurology, and cardiology \cite{sweet:1951:uses, ming:2020:progress}.
While providing unique in vivo molecular insights via radioactive tracers, PET imaging is inherently constrained by inferior image quality and signal-to-noise ratio (SNR) compared to structural modalities such as CT or MRI \cite{phelps2004molecular}.
This limitation is exacerbated by the clinical imperative to minimize radiation exposure and maximize patient throughput. 
Specifically, reducing tracer dosage or shortening acquisition times substantially degrades photon count statistics, elevating noise levels and compromising both lesion detectability and quantitative accuracy \cite{el2011improvement}. 
Consequently, developing advanced denoising techniques for low-dose PET remains important for supporting diagnostic confidence while reducing radiation exposure.

Traditionally, research has concentrated on conventional post-processing algorithms to enhance PET image quality.
For instance, standard spatial filters like Gaussian smoothing can reduce noise but may suppress high-frequency structural details, resulting in blurred fine structures.
To address this limitation and enable better preservation of detail, more advanced techniques, such as Nonlocal Mean \cite{dutta2013non}, wavelet methods \cite{shidahara2007pet}, guided filtering \cite{hashimoto2018denoising}, and block-matching filters \cite{ote2020kinetics} have been proposed.

More recently, the paradigm of PET image restoration has shifted towards deep learning-based solutions, propelled by the availability of extensive computational resources and large-scale datasets. 
These data-driven approaches have often reported improved noise suppression and structure preservation compared with traditional algorithms \cite{suzuki2017overview}.
Specifically, Convolutional Neural Networks (CNNs) have emerged as the backbone of modern PET denoising frameworks.
Among these, the U-Net architecture \cite{ronneberger2015u}, characterized by its encoder-decoder structure and skip connections, has been widely used \cite{jaudet2021impact, geng2021content, zhu2018image}. 
Despite its prevalence, the pixel-wise optimization objective in standard CNNs often leads to over-smoothed reconstructions, risking the obfuscation of small lesions or low-contrast textures.
To address the common issue of over-smoothing in U-Net-based denoising and enhance perceptual quality, Generative Adversarial Networks (GANs) have been explored. 
By introducing an adversarial loss that encourages the generator to produce outputs that are difficult to distinguish from real images, GANs can help preserve finer details \cite{zhou2020supervised, fu2023aigan}. 
However, their application is frequently hampered by training instability, sensitivity to hyperparameter tuning, and the inherent risk of mode collapse.

Denoising Diffusion Probabilistic Models (DDPMs) \cite{ho2020denoising} have recently achieved strong performance in high-fidelity generative modeling \cite{yu2025robust}. 
However, their clinical translation can be hindered by computational inefficiency. 
Specifically, the inference process entails numerically solving a stochastic differential equation (SDE) via thousands of sequential neural network evaluations, rendering the procedure slow for time-sensitive clinical workflows. 
Although acceleration algorithms such as Denoising Diffusion Implicit Models (DDIMs) \cite{song2020denoising} have been employed to expedite sampling, they commonly accelerate inference by reducing integration steps. 
This aggressive step reduction can introduce discretization errors, leading to a degradation in reconstruction fidelity and the loss of subtle structural details.

Beyond the computational bottlenecks discussed above, an important challenge for PET image denoising lies in generalization across the heterogeneity of clinical data, arising from varying tracers, dose levels, and scanner characteristics.
Deterministic models like CNNs typically learn a direct mapping from noisy to clean images; consequently, they may be sensitive to dataset-specific noise patterns and unseen distributions.
Similarly, adversarial approaches can be constrained by the sensitivity of the discriminator to domain shifts, which may lead to training instability or the generation of hallucinations on out-of-distribution data.
Even DDPMs, despite their strong generative capabilities, face a limitation beyond latency: their predefined noising process (typically Gaussian) may not fully adapt to PET-specific data characteristics, resulting in curved and potentially inefficient generative trajectories.
This landscape highlights the need for a framework that is both computationally efficient and flexible enough to navigate the complex manifold of PET data.

Rectified flow \cite{liu2022rectified} has recently emerged as a compelling generative framework grounded in Ordinary Differential Equations (ODEs).
Conceptually, rectified flow trains a velocity field on linear interpolants between samples from a source distribution and a target distribution.
In this formulation, the straight path is the ideal training target, and the learned ODE trajectory may deviate from this ideal in practice.
This direct transport formulation can enhance sampling efficiency, and rectified flow has demonstrated strong performance in natural image tasks, such as super-resolution and restoration, achieving high fidelity with fewer sampling steps compared to traditional diffusion models \cite{zhu:2024:flowie, qin:2025:reversing, ohayon:2025:posteriormean}.

However, the clinical translation of this framework faces two important challenges: numerical precision and dimensional limitations.
First, solving the learned ODE involves a delicate balance between discretization errors and accumulation errors. \cite{song2023consistencymodels, salimans2022progressivedistillationfastsampling, kim2024simplereflowimprovedtechniques}
Using too few steps introduces significant discretization bias, whereas an excessive number of steps can lead to the accumulation of truncation errors and network estimation inaccuracies, paradoxically degrading image quality despite higher computational costs.
Standard uniform sampling strategies are often suboptimal as they distribute computational budget evenly across the trajectory, even though the learned vector field may vary across time.
This inefficiency necessitates a higher number of steps to maintain quality, diminishing the theoretical speed advantage of the flow-based model.

Second, while preliminary studies have applied rectified flow to medical imaging \cite{sun2025pet, ma2025newoneshotfederatedlearning, yazdani2025flowmatchingmedicalimage}, most remain confined to the 2D domain.
Since PET is inherently volumetric, relying solely on 2D slice-wise operations does not explicitly model full 3D inter-slice spatial correlations.
This neglect often results in z-axis inconsistencies, where reconstructed volumes exhibit jagged discontinuities or fail to accurately capture the 3D morphology of lesions.
Such artifacts are particularly relevant in PET, where volumetric quantification and lesion conspicuity across all dimensions are important for clinical interpretation.

To address these limitations, we introduce and validate a novel 3D denoising framework for PET imaging based on rectified flow. 
This framework is explicitly designed to reconcile the trade-off between computational efficiency and reconstruction robustness. The primary contributions of this study are summarized as follows:

\begin{enumerate}
\item A One-Pass Conditional 3D Rectified Flow Framework Tailored for Whole-Body PET Denoising:
We adapt the rectified flow paradigm to whole-body PET imaging using a one-pass linear-interpolant conditional velocity-matching objective.
By modeling the joint distribution of full volumetric data, the proposed framework effectively leverages crucial inter-slice spatial contexts.
This approach is intended to reduce z-axis inconsistencies observed in 2D methods and to improve volumetric coherence and preservation of 3D anatomical structures and lesion integrity.

\item An Optimized Non-Uniform Sampling Strategy for Quality-Preserving and Rapid Inference:
We propose an empirically optimized fixed non-uniform sampling strategy that jointly selects the number of integration steps and the global time-stepping schedule for the conditional flow ODE solver.
The schedule is selected through ablation experiments and interpreted with trajectory-error analysis.
This dual optimization enables the 3D Flow model to achieve high reconstruction quality for both global metrics and local lesion features in approximately 30 seconds in our implementation, compared with multi-hour inference for the evaluated 3D DDPM baseline.

\item Zero-Shot Transfer across Unseen Doses and Datasets:
The proposed framework exhibits promising performance across the evaluated data distributions, spanning both public benchmarks and in-house clinical acquisitions.
Specifically, this is established through a strict zero-shot protocol: the model is trained exclusively on a single dose level from one dataset and directly applied to unseen datasets and a wide spectrum of dose reduction factors without any fine-tuning.
This suggests reduced sensitivity to dataset-specific noise patterns or scanning parameters within the evaluated settings, although broader validation remains necessary.
\end{enumerate}

\section{Methods and Materials}
\subsection{3D Rectified Flow for PET Image Denoising}
\begin{figure*}[!t]
    \centering
    \includegraphics[width=0.95\textwidth]{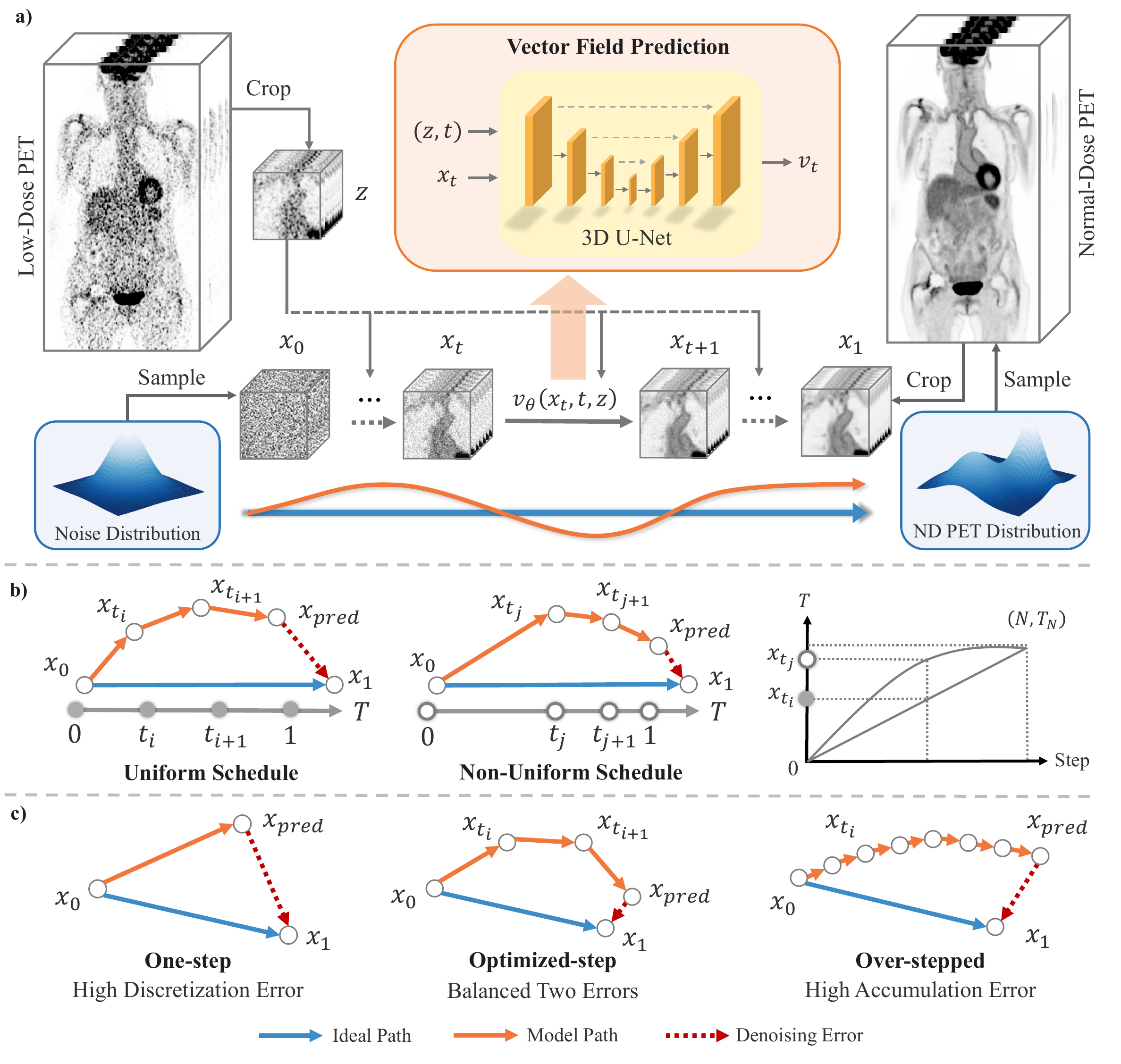}
    \caption{
        Overview of the proposed 3D rectified flow denoising framework with an optimized sampling strategy.
        {\bf{a)}} Model Workflow:
        Our framework utilizes a 3D U-Net to predict a conditional vector field $v_\theta(x_t, t, z)$, which transports the initial noise ($x_0$) to the target high-quality PET image ($x_1$), guided by low-dose input ($z$). 
        Note the deviation between the theoretical straight path (blue arrow) and the actual learned trajectory (orange curve).    
        {\bf{b)}} Schedule optimization: 
        The standard uniform sampling schedule (left) is contrasted with the proposed non-uniform schedule (middle).
        As schematically illustrated by the trajectory curve (right), the non-uniform strategy adjusts the density of time steps along the trajectory and is designed to allocate more integration resolution to selected portions of the path.
        {\bf{c)}} Denoising errors with varying steps:
        The diagram depicts the distinct error mechanisms observed at varying step counts.
        A one-step approach (left) suffers from high discretization error due to trajectory curvature.
        Conversely, an over-stepped approach (right) may accumulate small prediction or integration errors over repeated updates, leading to gradual endpoint drift while generally progressing toward the target.
        The optimized strategy (center) is designed to balance these two opposing error sources.
    }
    \label{fig:framework_and_sampling}
\end{figure*}

We propose a specialized 3D rectified flow framework explicitly designed for volumetric PET image denoising. 
The overall architecture, key conceptual challenges, and the associated sampling strategy are schematically illustrated in Fig.~\ref{fig:framework_and_sampling}.
Grounded in the theory of Ordinary Differential Equations (ODEs), our method formulates the denoising task as a deterministic transport problem.
Specifically, the model learns a continuous vector field that defines a transport trajectory from a tractable prior distribution (e.g., standard Gaussian noise) to the reference-dose PET data distribution.
During inference, by numerically integrating this learned ODE conditioned on the low-dose PET input, the model deterministically reconstructs a high-quality volumetric image from a random noise sample.
In this work, the term 3D Flow refers to a one-pass conditional rectified-flow formulation trained with a linear-interpolant velocity-matching objective.

\subsubsection{Conditional Rectified Flow Framework}
The task of PET image denoising is formally framed as a conditional generative modeling problem. 
Let $z \in \mathbb{R}^{H \times W \times D}$ denote a 3D low-dose PET image, which serves as the condition, and let $x_1 \in \mathbb{R}^{H \times W \times D}$ be the corresponding high-quality target image sampled from the normal-dose PET image distribution $p_1(x)$. 
Our goal is to learn a generative process to produce an accurate estimate, $\hat{x}_1$, from a tractable prior distribution, such as a standard Gaussian distribution $\mathcal{N}(0, I)$, guided by the conditional information from $z$.

We model this generative process with a conditional ODE with a time variable $t \in [0,1]$, where $t=0$ corresponds to the prior domain and $t=1$ corresponds to the normal-dose PET image domain. 
A continuous trajectory of images $x_t$ evolves from the prior distribution according to the ODE \cite{lipman2022flow}:
\begin{equation}
    \frac{dx_t}{dt} = v(x_t, t, z).
\end{equation}
Here, $x_t$ represents the 3D PET image state at time $t$, and $v(x_t, t, z)$ denotes the conditional vector field approximated by the 3D U-Net.

In our one-pass conditional formulation, the training target is constructed using a linear interpolant between a Gaussian noise sample and the paired reference-dose PET image.\cite{liu2022rectified}.
To implement this, we couple a random noise sample $x_0 \sim \mathcal{N}(0,I)$ (representing the state at $t=0$) with a clean target image $x_1$ (the state at $t=1$). 
The path $x_t$ is then defined as their linear interpolation:
\begin{equation}
    x_t = (1-t)x_0 + tx_1.
\end{equation}
By differentiating this path with respect to time $t$, we derive the ideal target vector field for this straight trajectory:
\begin{equation}
    \frac{dx_t}{dt} = x_1 - x_0.
\end{equation}
This equation defines the ideal constant velocity vector required to transport a random noise sample $x_0$ directly to a specific clean image $x_1$ along a linear path.
During inference, the learned ODE trajectory is therefore treated as an approximation to this ideal straight path.

\subsubsection{Model Training and Inference}
The neural network model $v_{\theta}$, parameterized by $\theta$, is trained to approximate the ideal vector field. 
The network is designed to predict the target instantaneous velocity $(x_1-x_0)$ given an intermediate image $x_t$, the time step $t$, and the conditional low-dose image $z$. 
The optimization objective is to minimize the Mean Squared Error (MSE) loss between the network prediction and the ground truth, formulated as:
\begin{equation}
    \mathcal{L}(\theta) = \mathbb{E}_{t,x_0,(z,x_1)}[\|v_{\theta}((1-t)x_0+tx_1, t, z) - (x_1-x_0)\|^2].
\end{equation}
In each training iteration, we sample a time $t$, a noise vector $x_0$, and a data pair $(x_1, z)$ to update the network parameters $\theta$.
This objective is optimized once on the original noise-target pairs.
The training procedure is summarized in Algorithm ~\ref{alg:training}. 

\begin{algorithm}[!h]
    \caption{Training the Conditional Vector Field}
    \label{alg:training}
    \begin{algorithmic}[1]
        \REPEAT
            \STATE Sample $t \sim U(0,1)$
            \STATE Sample $x_0 \sim \mathcal{N}(0,I)$, $(x_1, z) \sim p_{joint}(x_1, z)$
            \STATE Construct interpolated image: $x_t \gets (1-t)x_0 + tx_1$
            \STATE Compute loss: $\mathcal{L} \gets \|v_{\theta}(x_t, t, z) - (x_1-x_0)\|^2$
            \STATE Update parameters $\theta$ using gradient descent on $\mathcal{L}$
        \UNTIL{converged}
    \end{algorithmic}
\end{algorithm}

Once the vector field $v_{\theta}$ is learned, the model performs inference to reconstruct high-quality volumes from an unseen low-dose PET image $z$. 
The process, detailed in Algorithm~\ref{alg:sampling}, initiates by drawing a random Gaussian noise sample $x_0 \sim \mathcal{N}(0,I)$, which serves as the initial state $x_{t_0}$. 
We solve the learned ODE from $t=0$ to $t=1$ using a numerical solver over a sequence of $N$ discrete time steps, $\{t_0, t_1, \dots, t_N\}$, defined by a specific (potentially non-uniform) time schedule where $t_0=0$ and $t_N=1$. 
This is achieved by iterating from $i=0$ to $N-1$ and applying the Euler integration step for each interval $[t_i, t_{i+1}]$:
\begin{equation}
    x_{t_{i+1}} = x_{t_{i}} + (t_{i+1}-t_{i}) \cdot v_{\theta}(x_{t_{i}}, t_{i}, z).
\end{equation}
The process is initialized with $x_{t_0} = x_0$. 
The linear-interpolant training target encourages a direct transport field, which can support efficient few-step inference, although the learned trajectory remains an approximation.

\begin{algorithm}[!h]
    \caption{Sampling (Denoising)}
    \label{alg:sampling}
    \begin{algorithmic}[1]
        \STATE \textbf{Input:} Low-dose PET image $z$, number of steps $N$
        \STATE Initialize with noise: $x_{t_0} \gets x_0 \sim \mathcal{N}(0,I)$
        \STATE Time schedule function $T: \{0, \dots, N\} \to [0,1]$
        \FOR{$i=0, \dots, N-1$}
            \STATE Set current time: $t_i \gets T(i)$
            \STATE Compute next time: $t_{i+1} \gets T(i+1)$ 
            \STATE Compute step size: $dt \gets t_{i+1} - t_i$
            \STATE Compute velocity: $v_t \gets v_{\theta}(x_{t_i}, t_i, z)$
            \STATE Perform Euler step: $x_{t_{i+1}} \gets x_{t_i} + v_t \cdot dt$
        \ENDFOR
        \STATE \textbf{Return} Denoised image $\hat{x}_1 \gets x_{t_N}$
    \end{algorithmic}
\end{algorithm}

\subsubsection{Optimized Sampling Strategy for PET Image Denoising}
While rectified flow aims for an ideal straight-line trajectory, in practice, the neural network $v_{\theta}$ provides only an approximation of this ideal straight path, leading to subtle curvatures in the learned trajectory as illustrated in Fig.~\ref{fig:framework_and_sampling}b. 
When numerically integrating this ODE during inference (i.e., solving for $x_t$ from $t=0$ to $t=1$), the continuous path must be discretized into a finite sequence of steps. 
This discretization introduces approximation errors that may accumulate and affect final PET image quality.
The magnitude of this error is influenced by both the time-stepping schedule (distribution of steps) and the total number of sampling steps ($N$).

The choice of time-stepping schedule $\{t_0, t_1, \dots, t_N\}$ is therefore important. 
A uniform sampling schedule, where steps are evenly spaced across $t \in [0,1]$, might not be optimal. 
This is because the curvature of the learned trajectory $v_{\theta}(x_t, t, z)$ can vary significantly along $t$. 
As qualitatively suggested in Fig.~\ref{fig:framework_and_sampling}b, a uniform step size may over-sample some portions of the trajectory and under-sample others where the learned vector field changes more rapidly.
To address this limitation, we propose an optimized non-uniform sampling strategy.
This strategy uses a fixed parameterized time-stepping rule to bias the placement of sampling steps across the time domain.
The schedule is selected empirically through the ablation study described below and is used as a global time-stepping rule during inference.
The empirical support for this design is provided by the multi-dose schedule ablation and the trajectory-error analysis presented later.

\begin{figure}[!h]
\centering
\includegraphics[width=1.0\columnwidth]{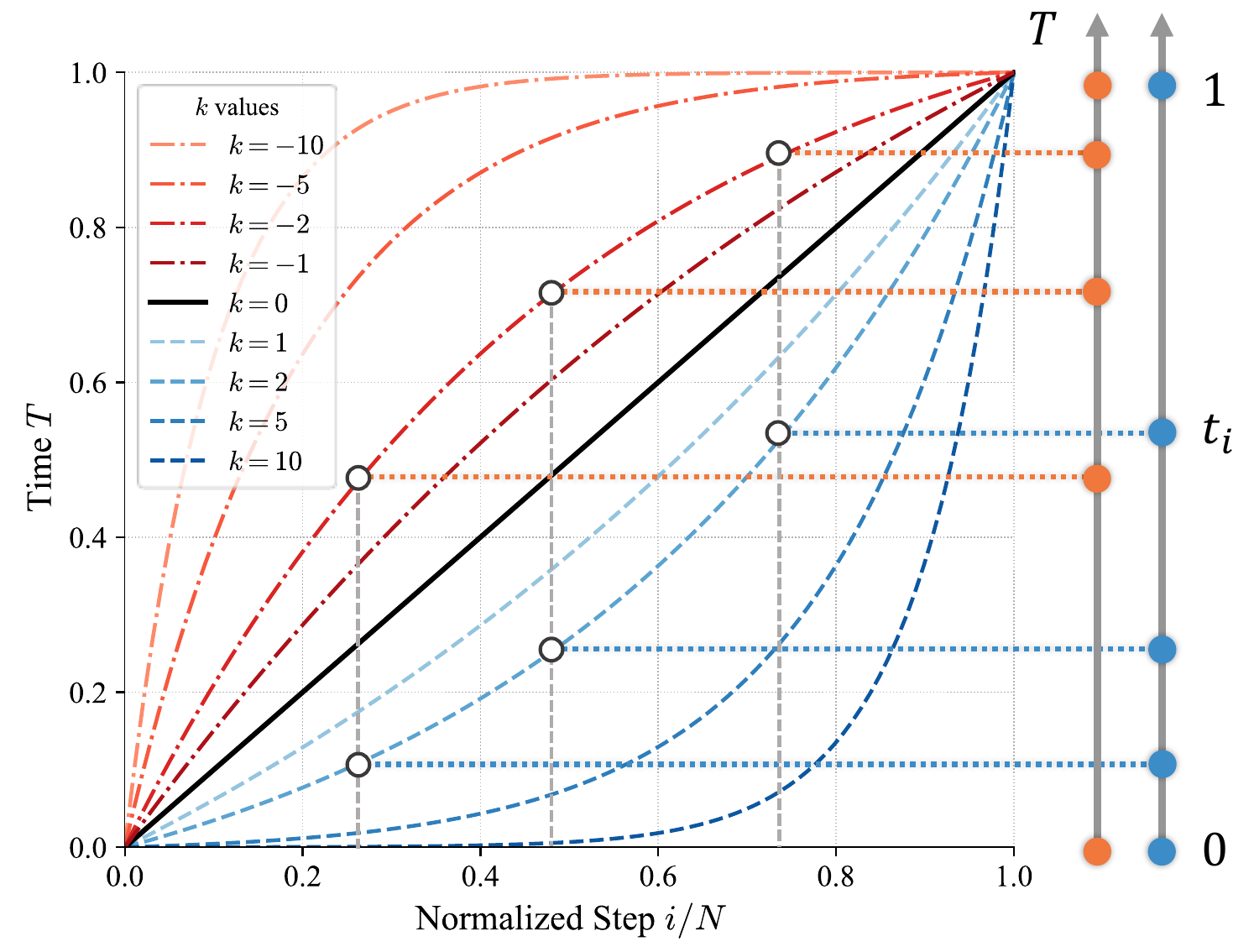}
\caption{
    Illustration of non-uniform sampling schedules controlled by the hyperparameter $k$.
    This figure shows the distribution of time steps $t_i$ as a function of the normalized step index $i/N$ for various values of the hyperparameter $k$.
}
\label{fig:sampling_schedules}
\end{figure}

Specifically, we employ a flexible non-uniform time-stepping schedule defined by an exponential function.
For a total of $N$ steps, the $i$-th time step $t_i \in [0, 1]$, where $i=0, \dots, N$, is determined by the following transformation:
\begin{equation}
    t_i = \frac{e^{k \cdot \frac{i}{N}} - 1}{e^k - 1}.
\end{equation}
Here, $k$ is a hyperparameter that controls the non-uniformity of the sampling distribution.
Fig.~\ref{fig:sampling_schedules} illustrates the behavior of this function for varying $k$ values:
\begin{itemize}
    \item When $k\to 0$, the function converges to the standard uniform (linear) schedule, $t_i = i/N$, corresponding to evenly spaced time steps (black solid line).
    \item For $k>0$, the schedule concentrates more steps toward the early part of the trajectory ($t=0$) (blue dashed lines).
    \item Conversely, for $k<0$, steps are concentrated toward the endpoint of the trajectory ($t=1$) (red dash-dotted lines).
\end{itemize}

Beyond the distribution of steps, the total number of sampling steps ($N$) also presents a critical trade-off for PET image denoising. 
While using a very small number of steps, such as a single step ($N=1$), offers maximum inference speed, the large step size can lead to discretization error, resulting in an output image that deviates from the ground truth image, as depicted in Fig.~\ref{fig:framework_and_sampling}c (left panel). 
Conversely, an excessive increase in $N$ reduces individual step errors but leads to higher cumulative error across steps and drastically prolongs inference time, thereby negating the efficiency advantage of rectified flow.
Thus, identifying an appropriate number of steps is essential for balancing fidelity and efficiency.

In summary, our approach optimizes the rectified flow inference for PET image denoising by combining an optimized non-uniform time-stepping schedule (controlled by $k$) with a carefully selected number of sampling steps ($N$). 
This combined strategy aims to accurately approximate the learned trajectory with minimal steps, thereby balancing reconstruction fidelity with efficient inference.

\subsection{Datasets}
Our study utilized two distinct 3D PET datasets for model training and evaluation: the public Ultra-low Dose PET Imaging Challenge (UDPET) dataset and an independent clinical dataset from the First Affiliated Hospital of Zhejiang University School of Medicine (FAHZU).
To facilitate a reliable evaluation of dose-related noise effects, all reconstructed image volumes in both cohorts were converted to Standardized Uptake Value (SUV) units.

\subsubsection{UDPET Dataset}
The Ultra-low Dose PET Imaging Challenge (UDPET) dataset \cite{Kuangyu2022Ultra} served as our primary resource for training and benchmarking. 
It comprises 377 whole-body $^{18}$F-FDG PET scans.
To ensure high image quality, the provided volumes were reconstructed using the ordered subset expectation maximization (OSEM) algorithm with 4 iterations and 5 subsets. 
The reconstruction also incorporated time-of-flight (TOF) and point-spread-function (PSF) modeling.
Subsequently, a post-reconstruction 2 mm full-width at half-maximum (FWHM) Gaussian filter was applied to further suppress noise \cite{alberts2021clinical}.
This dataset provides paired data covering a wide dynamic range. 
For each subject, a normal-dose reference is available alongside six simulated low-dose volumes, corresponding to  dose reduction factors (DRFs) of 2, 4, 10, 20, 50, and 100.
All volumes feature an isotropic voxel size of $1.65 \times 1.65 \times 1.65$ mm$^3$.

\subsubsection{FAHZU Dataset}
To assess the model's robustness on independent clinical data, we utilized a retrospective collection from the First Affiliated Hospital of Zhejiang University School of Medicine (FAHZU).
Ethical approval was granted by the institutional Ethics Committee (Approval No. [2025B]IIT-0189). 
The requirement for informed consent was waived due to the retrospective nature of the study.
This cohort includes 50 patients featuring clinically verified lesion annotations.
The lesion annotations included 519 lesions in total, with a median of 9.5 lesions per patient (interquartile range 4.75--14.0; range 1--36).
The median lesion diameter was 0.7 cm (interquartile range 0.5--1.1 cm; range 0.3--12.0 cm).
The dataset was acquired at a different clinical center on a scanner recorded as Biograph64 Vision 600.
Images were reconstructed using OSEM with 4 iterations and 5 subsets, with TOF and PSF modeling.
Following reconstruction, a 2 mm FWHM Gaussian filter was applied.
For each patient, a standard normal-dose image was acquired with a duration of 240 seconds per bed. 
The low-dose inputs were derived from respiratory-gated acquisitions triggered at end-exhalation (using a 35\% phase width).
Since only a portion of the respiratory cycle was used, the effective scan time was reduced. Specifically, while the nominal acquisition times were 30, 60, 90, and 120 seconds per bed (s/bed), the effective times used for reconstruction were approximately 10, 20, 30, and 40 s/bed, respectively.

Compared with UDPET, the FAHZU evaluation therefore represents an independent-center, protocol-level domain shift.
The two cohorts differ in cohort source, scanner model, low-dose generation mechanism, acquisition duration, and the availability of voxel-size metadata, while the available reconstruction information indicates aligned OSEM settings with TOF/PSF modeling and 2 mm Gaussian post-filtering.

\subsection{Evaluation Metrics}
The performance of our denoising framework was quantitatively evaluated using metrics designed to assess both global image fidelity and local lesion reconstruction quality.

\subsubsection{Global Image Quality Assessment}
To assess global image fidelity, we employed two widely used full-reference metrics: the Peak Signal-to-Noise Ratio (PSNR) and the Structural Similarity Index Measure (SSIM) \cite{wang2004image}.

First, PSNR quantifies reconstruction quality by measuring the ratio of a signal's maximum possible power to the power of corrupting noise. 
In the context of PET, a higher PSNR value indicates effective suppression of the inherent statistical noise while maintaining fidelity to the ground-truth Standardized Uptake Value (SUV) range. 
It is calculated as:
\begin{equation}
    \text{PSNR}(x, y) = 20 \cdot \log_{10}\left(\frac{\text{MAX}_{SUV}}{\sqrt{\text{MSE}(x,y)}}\right),
\end{equation}
where $\text{MAX}_{SUV}$ is the maximum possible SUV in the normal-dose PET images, and MSE is the mean squared error.

Second, SSIM provides a complementary assessment by evaluating the perceptual similarity between two images based on three components: luminance, contrast, and structural information. 
These three components hold specific clinical significance in PET imaging analysis: luminance corresponds to the mean regional tracer uptake, contrast is critical for lesion conspicuity against background tissue, and structure relates to the preservation of anatomical boundaries. 
Therefore, a high SSIM score suggests that the denoised image is structurally closer to the reference image according to this metric. 
The SSIM is calculated as:
\begin{equation}
    \text{SSIM}(x, y) = \frac{(2\mu_x\mu_y + c_1)(2\sigma_{xy} + c_2)}{(\mu_x^2 + \mu_y^2 + c_1)(\sigma_x^2 + \sigma_y^2 + c_2)},
\end{equation}
where $\mu$ and $\sigma$ represent the mean and standard deviation of SUV, respectively, $\sigma_{xy}$ is the covariance, and $c_1, c_2$ are stabilizing constants. 

\subsubsection{Local Lesion Quantification}
Beyond global image fidelity, focal-lesion conspicuity is an important image-quality consideration in PET.
To assess local lesion conspicuity, we measured the Contrast-to-Noise Ratio (CNR) \cite{cui2019pet}.
In this study, CNR is used as a lesion-conspicuity metric that summarizes lesion-to-background contrast normalized by background variability.
It is formulated as:
\begin{equation}
    \text{CNR} = \frac{|\mu_{\text{lesion}}-\mu_{\text{ref}}|}{\sigma_{\text{ref}}},
\end{equation}
where $\mu_{\text{lesion}}$ represents the mean SUV within the lesion's region of interest (ROI), while $\mu_{\text{ref}}$ and $\sigma_{\text{ref}}$ are the mean and standard deviation of the SUV in a nearby background reference region, respectively.
Lesion ROIs were delineated on the 240 s normal-dose PET images with assistance from the corresponding CT images by two imaging physicians, each with more than two years of relevant experience.
For background reference measurement, the physicians selected an appropriate liver region and placed a spherical ROI with a diameter of 3 cm.
The same lesion and background ROIs were applied to the low-dose and denoised images for paired comparison across methods.

\subsubsection{Statistical Analysis}
To determine the statistical significance of the differences between our proposed method and the baseline models, we performed the Wilcoxon signed-rank test. 
This non-parametric test was chosen because the distribution of the metric scores across the patient cohort could not be assumed to be Gaussian, making it a more robust alternative than paired t-tests\cite{Wilcoxon1992}.
For families of multiple pairwise comparisons, Benjamini--Hochberg false discovery rate (FDR) correction was applied.

\begin{figure*}[!htp]
\centerline{\includegraphics[width=1.0\textwidth]{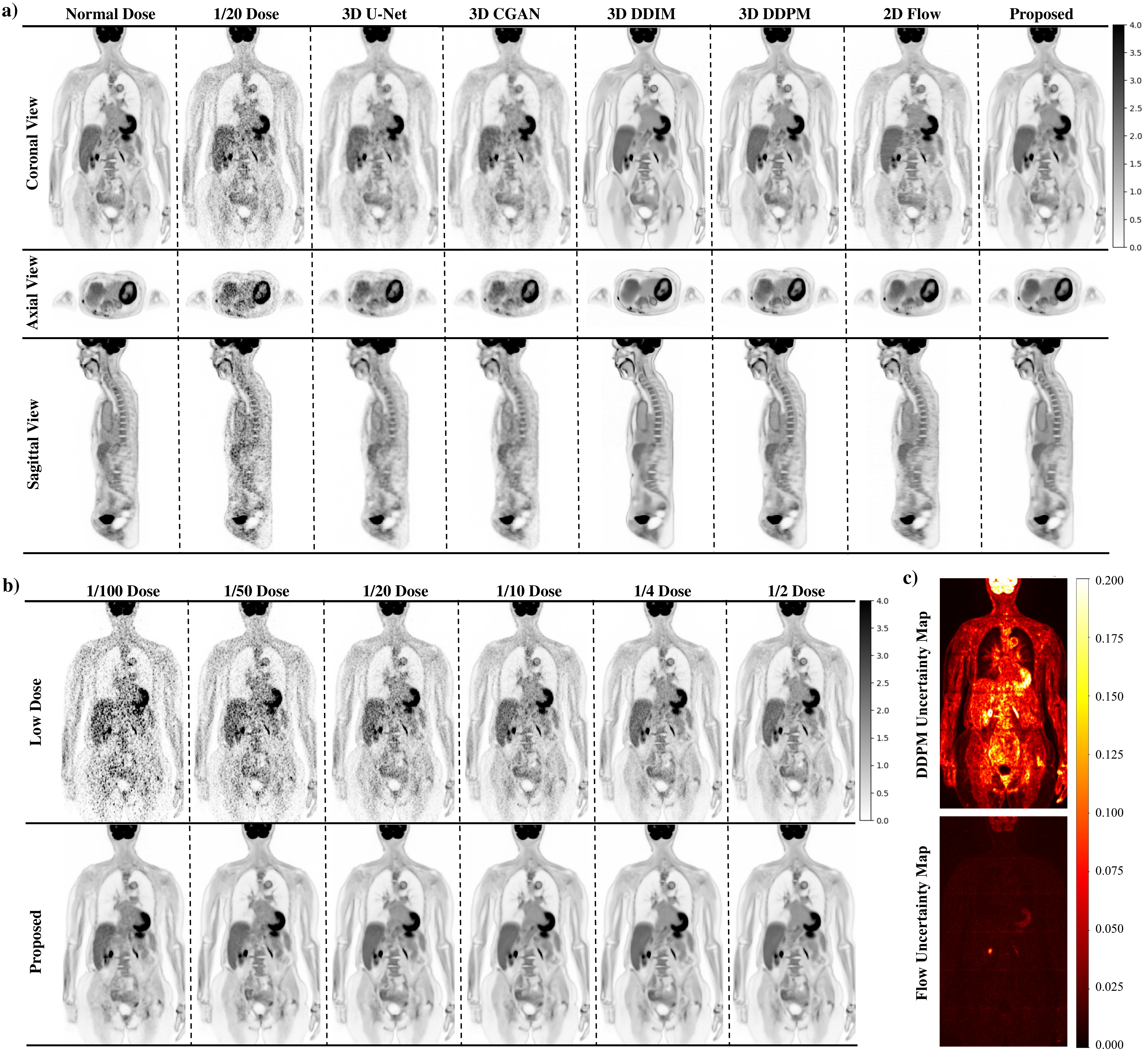}}
\caption{
    Denoising performance and reconstruction uncertainty on UDPET Dataset.
    {\bf{a)}} Qualitative comparison at matched 1/20 dose: Coronal, axial, and sagittal views comparing the 1/20 dose noisy input, its normal-dose ground truth, and the denoised outputs from various methods (3D U-Net, 3D CGAN, 3D DDIM (200 step), 3D DDPM, 2D Flow, Proposed).
    {\bf{b)}} Denoising across multiple dose levels: qualitative comparison of low-dose PET inputs (ranging from 1/100 to 1/2 doses) with the corresponding normal-dose image and the denoised output from our proposed 3D rectified flow method.
    {\bf{c)}} Reconstruction uncertainty maps: Standard deviation maps (calculated from 20 reconstructions with different random seeds) illustrating the per-voxel uncertainty of 3D DDPM (top) and our proposed 3D rectified flow (bottom) on a representative UDPET case.
}
\label{fig:udpet_over_models}
\end{figure*}

\begin{figure*}[!t]
\centerline{\includegraphics[width=1.0\textwidth]{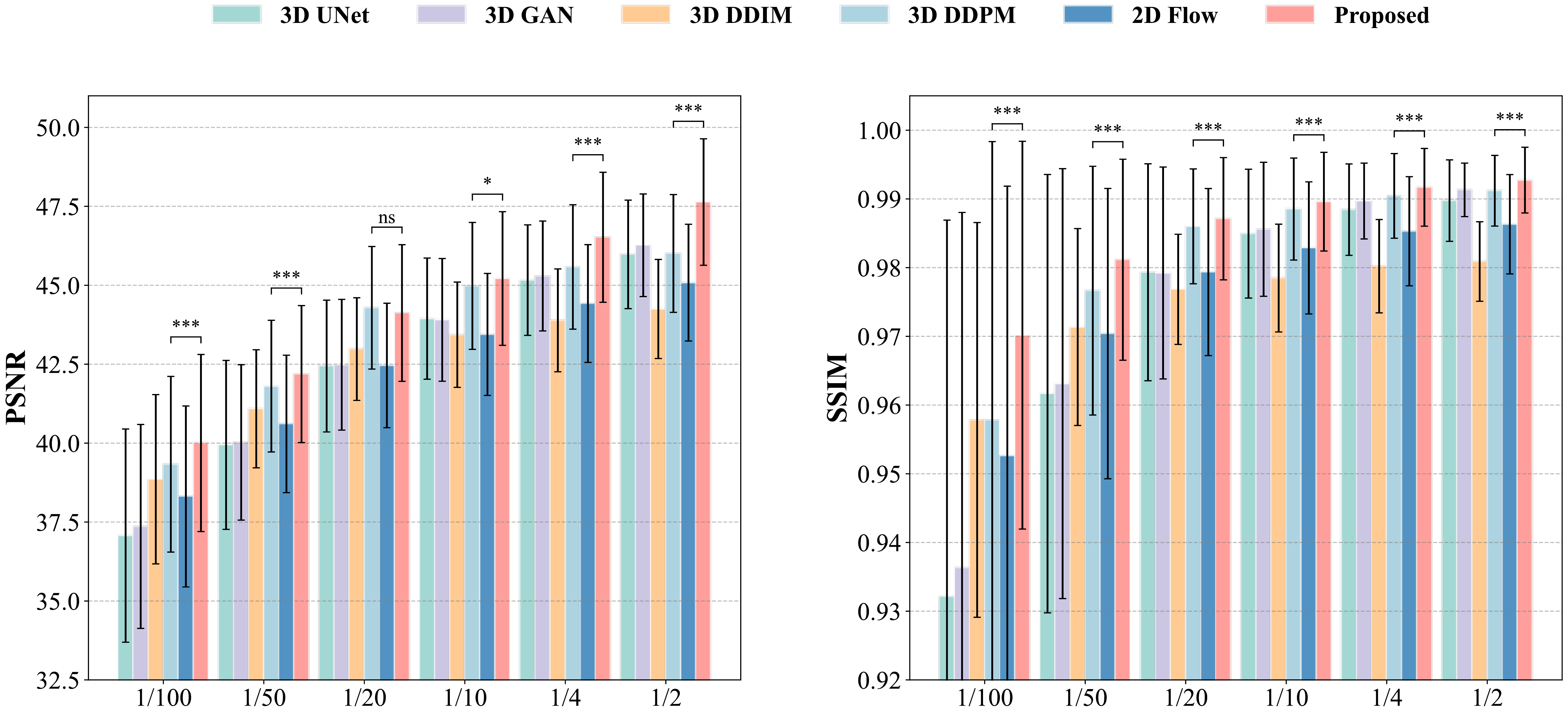}}
\caption{
    Quantitative performance on UDPET dataset.
    Bar plots showing PSNR and SSIM values (mean $\pm$ standard deviation) for different denoising methods across six low-dose levels (1/100 to 1/2) on the UDPET dataset.
    The proposed 3D rectified flow achieved favorable scores across the evaluated dose levels.
    Statistical significance (Wilcoxon signed-rank test against the next best baseline) is indicated: * for p $<$ 0.05, ** for p $<$ 0.01, and *** for p $<$ 0.001. ns denotes not significant.
}
\label{fig:udpet_metrics}
\end{figure*}

\begin{figure*}[!htbp]
\centerline{\includegraphics[width=1.0\textwidth]{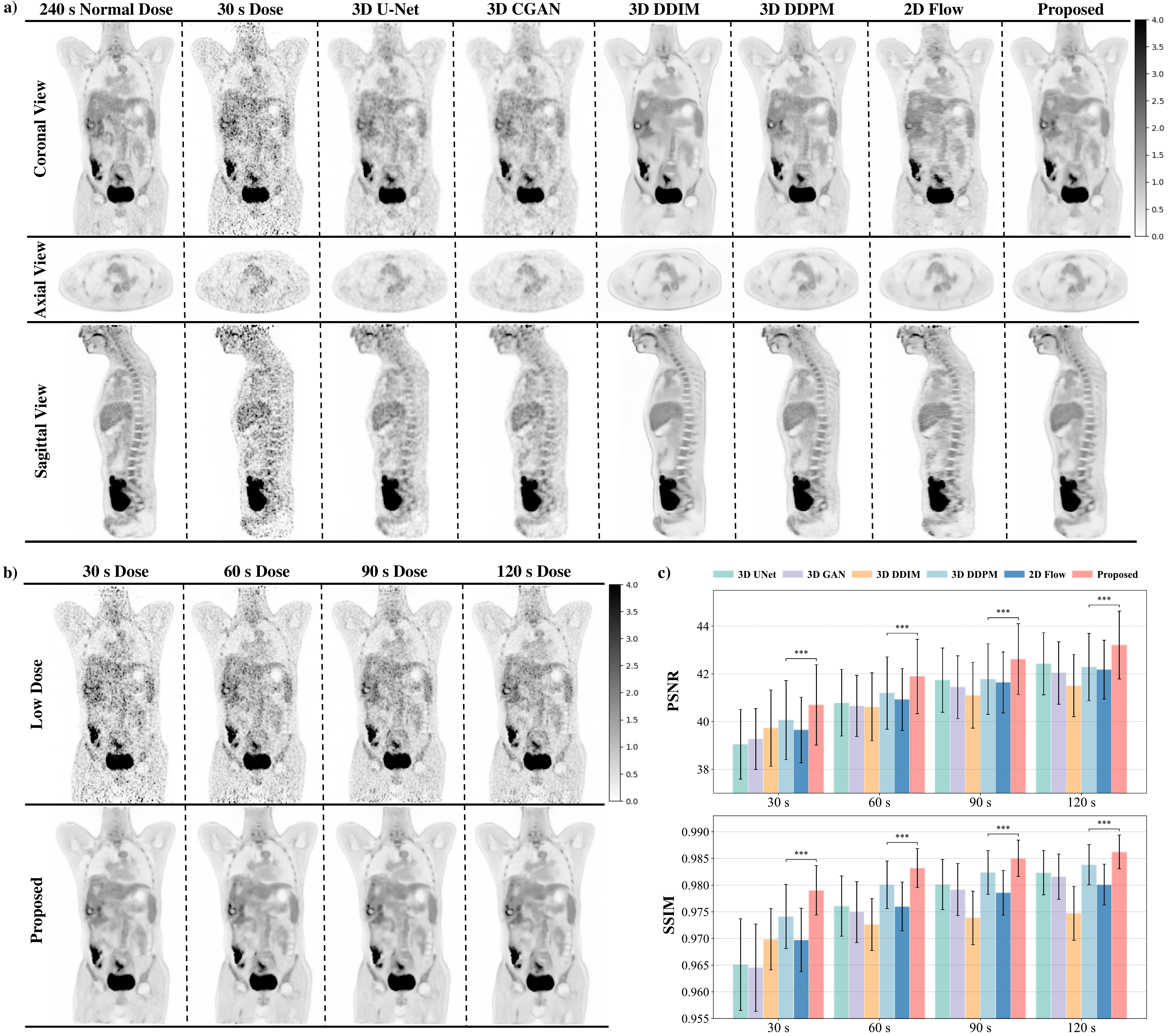}}
\caption{
    Denoising performance on FAHZU dataset.
    {\bf{a)}} Qualitative comparison at 30 s dose: Coronal, axial, and sagittal views comparing the 30 s dose noisy input, its 240 s normal-dose ground truth, and the denoised outputs from various methods (3D U-Net, 3D CGAN, 3D DDIM (200 step), 3D DDPM, 2D Flow, Proposed).
    {\bf{b)}} Denoising across multiple dose levels: Visual comparison of low-dose PET inputs (30 s, 60 s, 90 s, 120 s doses) with the 240 s normal-dose ground truth and the denoised output from our proposed 3D rectified flow method.
    {\bf{c)}} Quantitative performance: Bar plots showing PSNR and SSIM values (mean $\pm$ standard deviation) for different denoising methods across four dose levels (30 s, 60 s, 90 s, 120 s) on the FAHZU dataset.
    The proposed 3D rectified flow (Proposed) achieved favorable scores across the evaluated acquisition durations.
    Statistical significance (Wilcoxon signed-rank test against the next best baseline) is indicated: * for p $<$ 0.05, ** for p $<$ 0.01, and *** for p $<$ 0.001.
}
\label{fig:hzpet_over_models}
\end{figure*}

\begin{figure*}[!htbp]\centerline{\includegraphics[width=1.0\textwidth]{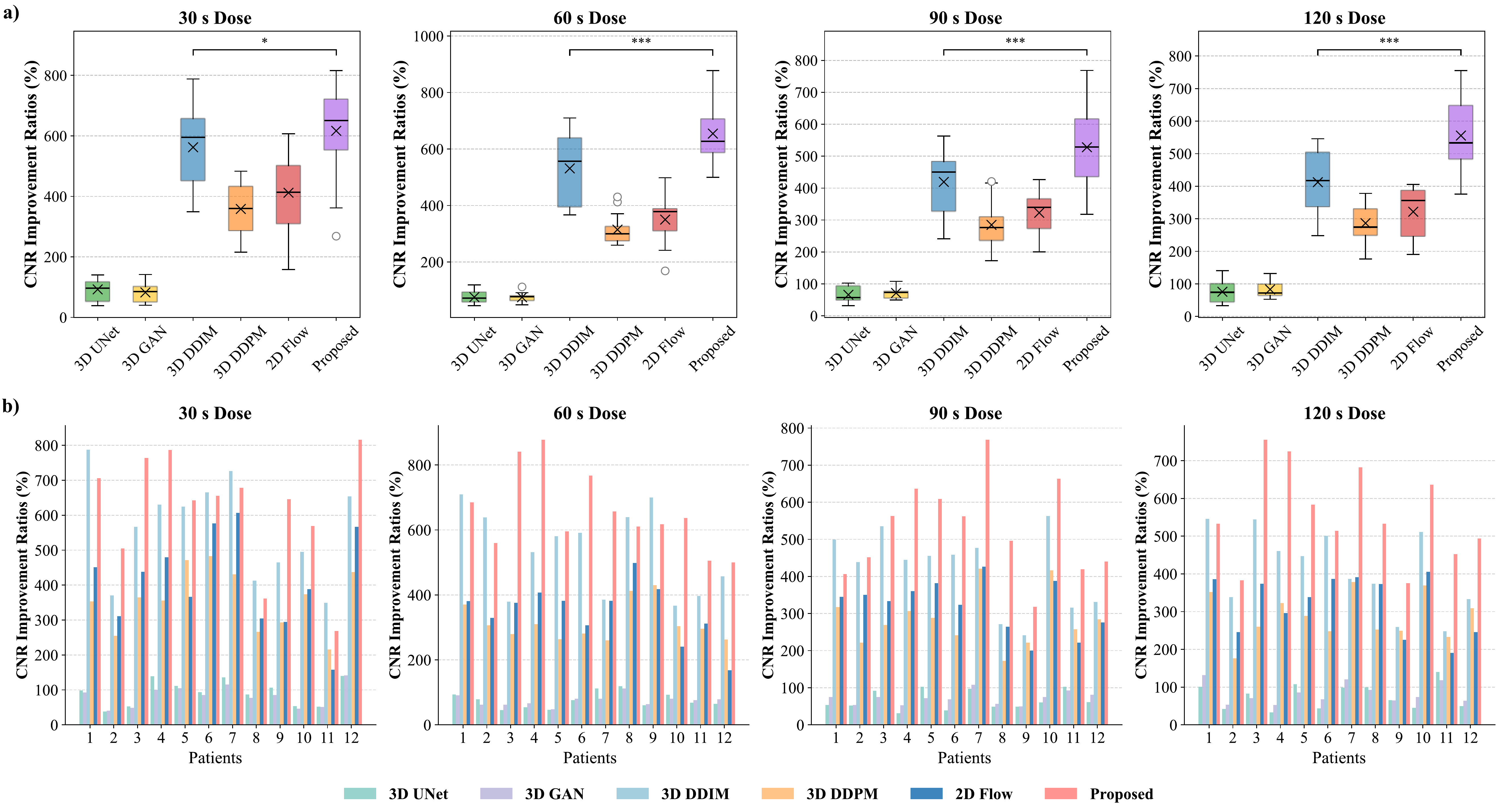}}
    \caption{
        Local lesion CNR improvement on FAHZU dataset.
        a) Overall CNR improvement ratios: Box plots illustrating the distribution of CNR improvement ratios for various denoising methods across four dose levels (30 s, 60 s, 90 s, 120 s) on the FAHZU dataset. 
        CNR is influenced by both lesion-background contrast and background variance and should be interpreted together with global image-quality metrics and visual assessment.
        Statistical significance (Wilcoxon signed-rank test against the next best baseline) is indicated: * for p $<$ 0.05, ** for p $<$ 0.01, and *** for p $<$ 0.001.
        b) Patient-specific CNR improvement ratios: Bar plots showing the CNR improvement ratios for 12 individual patients across different dose levels (30 s, 60 s, 90 s, 120 s) and various denoising methods.}
        \label{fig:FAHZU_CNR_results}
\end{figure*}

\subsection{Implementation Details}
\subsubsection{Experiment Setup and Data Partitioning}
The UDPET dataset was partitioned into dedicated training, validation, and internal test sets.
To evaluate the model's generalization capabilities, we employed a single-condition training strategy. 
Specifically, we utilized images corresponding exclusively to a single dose reduction factor (DRF) of 20 for both training (300 patients) and validation (16 patients) cohorts.
This specific dose level was selected because its moderate noise characteristics present a challenging learning target, intending to promote greater model robustness compared to training on extremely noisy or relatively clean data.

The trained model was subsequently evaluated on two distinct, unseen test cohorts without any fine-tuning:
\begin{itemize}
    \item Internal Test Set: The remaining 50 patients from the UDPET dataset, used to assess robustness across all six available DRFs (ranging from 100 to 2).
    \item External Test Set: The entire 50-patient FAHZU dataset with lesion annotations, used to evaluate domain generalization across different clinical conditions and acquisition times (ranging from 30 s to 120 s).
\end{itemize}

\subsubsection{Network Architecture and Baseline Models}
Our proposed denoising network is built upon a 3D U-Net architecture with 206,957,697 trainable parameters, reported as approximately 207 M.
For the 3D Flow model, the current state $x_t$ and the low-dose PET condition $z$ are concatenated along the channel dimension before the first 3D convolution.
The scalar time variable $t$ is encoded using a sinusoidal timestep embedding followed by a small multilayer projection, and the resulting embedding is injected into the residual blocks during velocity prediction.
To facilitate a rigorous comparative analysis, the main denoising and generative baselines used the same U-Net-based reconstruction backbone where applicable.
Representing standard deep learning paradigms, a standard 3D U-Net and a 3D GAN were included to serve as discriminative and adversarial baselines, respectively.
For the adversarial baseline, this capacity matching refers to the generator/reconstruction network; the discriminator is an auxiliary training component.

To specifically evaluate the impact of full 3D volumetric processing, we implemented a 2.5D version of our rectified flow model, hereinafter referred to as 2D Flow for comparisons. 
Guided by literature suggesting the advantages of 2.5D over pure 2D approaches \cite{xie2023ddpet3ddoseawarediffusionmodel}, this baseline processes a central slice along with two adjacent axial slices ($192 \times 288 \times 3$) to incorporate limited axial context. 
Furthermore, to benchmark against strong generative baselines, we compared our model with a standard 3D DDPM utilizing 1000 sampling steps and an accelerated 3D DDIM \cite{jiang2025fast} configured with 200 steps. 
The DDIM baseline was implemented using the reproduced baseline source code, and the 200-step setting was used as the main optimized DDIM configuration.
Lower-step DDIM settings were retained only for inference-efficiency analysis because they showed substantial degradation in this full-volume 3D PET denoising task.
We additionally evaluated DPM-Solver++ as a representative fast ODE-solver baseline applied to the reproduced 3D diffusion model.

Unlike the diffusion-based baselines, the number of ODE inference steps for our proposed 3D Flow model was set to only $N=2$.
This configuration was selected based on our ablation studies, which indicated that the proposed method achieves effective performance and high fidelity within this ultra-low step count range.

\subsubsection{Training and Inference Protocols}
Due to the large dimensionality of 3D PET volumes and GPU memory constraints, a specific data handling strategy was implemented. 
First, to remove non-essential background regions and standardize input size, all PET images were preprocessed by cropping them to a uniform volume of $192 \times 288 \times 520$ voxels. 
During the training phase, these cropped volumes were randomly partitioned into smaller $96 \times 96 \times 96$ patches with a stride of 48 voxels, facilitating memory-efficient learning.

For inference, a deterministic sliding-window approach was adopted. 
Each patient volume was divided into six $192 \times 288 \times 96$ patches along the axial direction, with a 10-voxel overlap between adjacent patches to mitigate edge artifacts. 
The final denoised volume was reconstructed by applying a weighted averaging scheme to these overlapping regions. 
During training, the initial noise $x_0$ was independently sampled from $\mathcal{N}(0,I)$ for each training sample and iteration.
During inference, random seeds were fixed to ensure reproducible evaluation and paired comparisons across sampling schedules.
The uncertainty analysis in Fig.~\ref{fig:udpet_over_models}c used 20 independent random seeds to assess sensitivity to noise initialization.

The 3D Flow model was implemented in PyTorch and trained using the AdamW optimizer. 
The learning rate followed a cosine annealing schedule, decaying from an initial value of $1\times10^{-4}$ to a minimum of $1\times10^{-8}$. 
Training was distributed across four NVIDIA RTX 4090 GPUs with a global batch size of 4 and converged after approximately 59 hours. 
To prevent overfitting, the checkpoint yielding the highest PSNR on the validation set was selected via early stopping. 
During the inference phase, the optimized model processes a full 3D patient volume in approximately 30 seconds on a single NVIDIA RTX 4090 GPU.
For runtime comparison, we used a standardized inference-only estimate under the same sliding-window protocol to avoid differences caused by data-loading implementations across baselines.

\section{Experiments and Results}
This section presents a quantitative and qualitative evaluation of the proposed 3D rectified flow framework for whole-body PET image denoising. 
The results are organized into three subsections to assess different aspects of model performance:

First, we evaluate the model's global image quality and generalizability across two multi-dose datasets. 
This analysis investigates the framework's performance under varying noise levels and its applicability to unseen clinical data without fine-tuning.

Second, we assess local lesion reconstruction quality. Given the importance of lesion conspicuity for diagnosis, we examine the model's effectiveness in preserving focal tracer uptake and enhancing contrast recovery relative to baseline methods.

Finally, we present an analysis of inference efficiency and an ablation study on the sampling strategy. 
This subsection investigates the trade-off between reconstruction fidelity and computational speed, aiming to validate the impact of the proposed non-uniform sampling schedule.

\subsection{Global Image Quality and Generalizability}
\subsubsection{Performance on Matched Training Conditions}
We first evaluated the model's performance on the UDPET test set at the 1/20 dose level, matching the noise level utilized during the training phase.
Qualitative results presented in Fig.~\ref{fig:udpet_over_models}a illustrate that our proposed 3D Flow model yields images with high clarity and anatomical fidelity, closely resembling the normal-dose ground truth.
Specifically, the model effectively suppresses the grainy background noise observed in the low-dose input without introducing artificial smoothing. 
In contrast, the standard U-Net baseline exhibits characteristic over-smoothing, resulting in a loss of fine textural details that obscures subtle structural features. 
The CGAN baseline, while retaining sharper edges, tends to introduce noticeable residual noise artifacts.
Notably, comparisons in the coronal and sagittal views highlight the benefit of native 3D processing. 
The 2D Flow model exhibits subtle z-axis inconsistencies, manifesting as streak-like artifacts across slices, whereas our 3D Flow model preserves seamless volumetric continuity.

Quantitatively (Fig.~\ref{fig:udpet_metrics}), at this matched 1/20 dose level, our 3D Flow model achieves performance comparable to the evaluated 3D DDPM baseline, yielding favorable PSNR and SSIM scores.
Beyond fidelity, reconstruction stability was assessed by analyzing the voxel-wise standard deviation across 20 distinct realizations (Fig.~\ref{fig:udpet_over_models}c).
The resulting uncertainty maps reveal a clear distinction. 
The 3D DDPM exhibits elevated variance (indicated by brighter regions), particularly localized around high-uptake structures such as the brain and lesions. 
In contrast, our 3D Flow model maintains consistently lower and more uniform uncertainty levels across the entire volume.
This enhanced deterministic stability implies that the flow-based method offers more reproducible outcomes, potentially reducing the risk of variability in clinical quantification.

\subsubsection{Robustness to Unseen Dose Levels}
We next investigated the model's zero-shot transfer to unseen noise levels, evaluating performance across a spectrum ranging from extremely low (1/100 dose) to relatively high (1/2 dose) counts.
As visualized in Fig.~\ref{fig:udpet_over_models}b, the 3D Flow model maintains denoising performance under extreme noise. In the 1/100 dose input, anatomical structures are largely submerged in severe statistical noise.
Notably, the model recovers the primary morphological features from this degraded input, improving the visibility of the skeletal structures and major organs that are barely discernible in the original low-dose image.

Quantitative analysis (Fig.~\ref{fig:udpet_metrics}) supports these qualitative observations. 
While the performance gap between methods narrows at higher doses (e.g., 1/2 dose), the advantage of the proposed method becomes increasingly pronounced as the dose decreases.
Crucially, in the challenging ultra-low-dose regimes (1/100 and 1/50), our 3D Flow model exhibits statistically significant improvements ($p<0.001$) over 3D DDPM.
This suggests that the conditional flow formulation and reduced sampling stochasticity may be beneficial in high-noise settings, although the learned trajectory should be interpreted as an approximation to the linear-interpolant training target.

\subsubsection{Zero-Shot Generalization to External Clinical Data}
To evaluate clinical generalizability, we applied the model, trained solely on UDPET data, directly to the independent FAHZU dataset without any fine-tuning.
This introduces a protocol-level domain shift between independent clinical centers, including differences in cohort source, scanner model, low-dose generation mechanism, and acquisition duration, ranging from 30 s to 120 s per bed (corresponding to effective durations of approx. 10 s to 40 s).

Qualitative results for the ultra-low-dose 30 s condition are shown in Fig.~\ref{fig:hzpet_over_models}a.
Under increased noise and apparent domain differences, the 3D Flow model yields images with preserved structural patterns and strong noise suppression compared with the normal 240 s dose reference.
The visual comparison also illustrates different noise-detail trade-offs among generative methods.
The 200-step DDIM output is generally smoother, whereas 3D DDPM may retain more high-frequency texture that can include both fine uptake detail and residual noise.

To further illustrate the model's generalization capability across the dose level, Fig.~\ref{fig:hzpet_over_models}b provides a visual comparison of reconstructions at varying acquisition times (30 s, 60 s, 90 s, and 120 s).
The 3D Flow model maintains consistent structural preservation and noise suppression across this range of dose levels.
This indicates that the model's zero-shot transferability is not limited to a specific noise magnitude within the evaluated clinical acquisition protocols of the external dataset.

Quantitatively (Fig.~\ref{fig:hzpet_over_models}c), our 3D Flow model yields statistically higher metrics than the evaluated baselines across all acquisition durations.
The advantage is also observed in the ultra-short 10 s and 20 s scenarios.

These findings support the potential for zero-shot transfer within the evaluated clinical protocols and noise distributions.

\begin{table*}[!htbp]
    \centering
    \caption{\textbf{Inference Speed and Performance Comparison on the FAHZU Dataset (30 s Dose)}}
    \label{tab:inference_speed}
    \begin{tabular}{lcccc}
        \toprule
        \textbf{Method} & \textbf{Steps} & \textbf{PSNR (dB)} & \textbf{SSIM} & \textbf{Inference Time} \\
        \midrule
        3D DDPM & 1000 & 40.06 $\pm$ 1.66 & 0.9741 $\pm$ 0.0060 & 4.27 h \\
        3D DDIM & 200 & 39.72 $\pm$ 1.60 & 0.9699 $\pm$ 0.0057 & 51.23 min \\
        3D DDIM & 100 & 25.19 $\pm$ 1.39 & 0.6369 $\pm$ 0.0383 & 25.62 min \\
        DPM-Solver++ & 100 & 39.22 $\pm$ 1.50 & 0.9269 $\pm$ 0.0151 & 25.62 min \\
        DPM-Solver++ & 50 & 28.99 $\pm$ 0.14 & 0.2397 $\pm$ 0.0225 & 12.81 min \\
        \textbf{Proposed (3D Flow)} & \textbf{2} & \textbf{40.70 $\pm$ 1.68} & \textbf{0.9790 $\pm$ 0.0046} & \textbf{30.74 s} \\
        \bottomrule
    \end{tabular}
\end{table*}

\subsection{Local Lesion Reconstruction Quality Evaluation}
Beyond global image fidelity, the reconstruction of focal tumor lesions is an important image-quality consideration.
In this section, we evaluate lesion conspicuity using the Contrast-to-Noise Ratio (CNR) improvement ratio on the FAHZU dataset, where lesion annotations are available.

The CNR improvement is defined as the percentage increase in CNR of the denoised image relative to the noisy low-dose PET input. 
By normalizing the improvement against the baseline low-dose input, this metric provides a summary of changes in lesion-to-background conspicuity.
Because CNR depends on both lesion-background contrast and the background standard deviation, we interpret it together with global image metrics and qualitative comparisons.

Fig.~\ref{fig:FAHZU_CNR_results}a presents box plots illustrating the distribution of CNR improvement ratios across all analyzed lesions. 
These distributions are stratified by acquisition duration (30 s, 60 s, 90 s, and 120 s) to demonstrate performance under varying noise conditions.

The distributions show that our proposed 3D Flow model achieves favorable median CNR improvement ratios across the evaluated dose levels.
Deterministic baselines such as U-Net show limited improvement, likely due to the suppression of focal uptake intensity during smoothing.
Diffusion-based methods can also improve CNR, but their CNR ranking should be interpreted cautiously because methods that strongly smooth the background can reduce $\sigma_{\text{ref}}$ and thereby increase CNR without necessarily improving structural fidelity.
Statistical analysis with FDR correction supports a significant difference between the proposed method and the evaluated baselines in the FAHZU lesion-conspicuity analysis.
We observe that the interquartile range (IQR) of the 3D Flow results indicates some variability. 
However, this spread may reflect natural biological heterogeneity of the lesions (varying in size, uptake, and location); additional analyses would be needed to separate lesion-related variability from model variability.

To provide a more granular insight into patient-specific performance, Fig.~\ref{fig:FAHZU_CNR_results}b showcases the CNR improvement ratios for lesions from 12 representative patients across all dose levels. 
Some baseline methods demonstrate competitive performance for specific lesions, and the proposed method shows generally favorable lesion-conspicuity changes across the representative cases.
Formal reader studies and lesion-level detection analysis are still needed to evaluate diagnostic impact.

Collectively, these results support the use of CNR improvement as an auxiliary lesion-conspicuity assessment.
They do not establish SUV quantitative equivalence or diagnostic performance, which require dedicated future validation.

\subsection{Inference Efficiency and Ablation Analysis}
\label{sec:ablation}
The practical utility of PET denoising models depends not only on image restoration quality but also on computational efficiency during inference.
This section benchmarks the inference speed of our proposed 3D Flow model against representative generative baselines and presents an ablation study to evaluate the optimization of our sampling strategy.

\begin{figure}[!hbp]
\centering
\includegraphics[width=1.0\columnwidth]{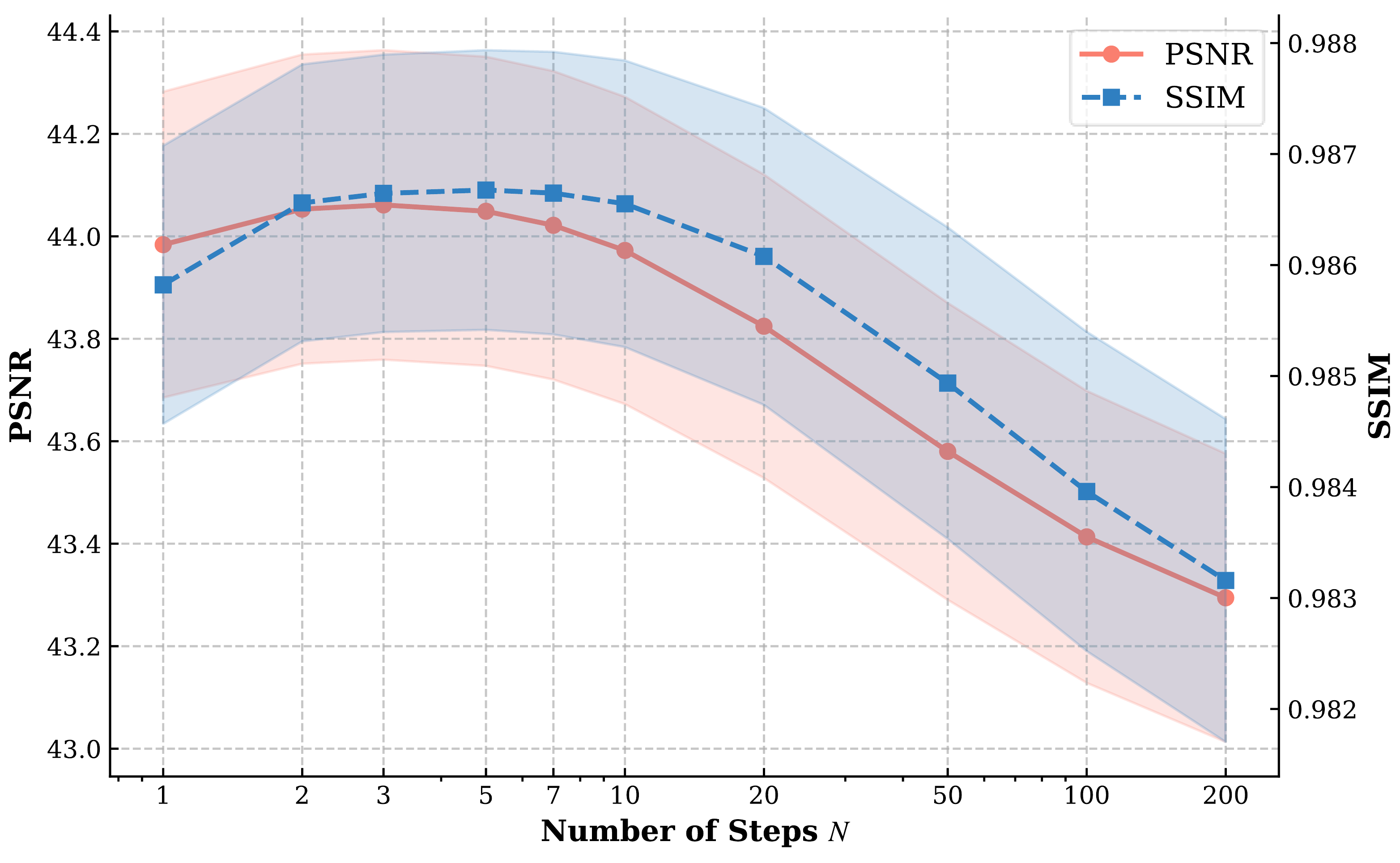}
\caption{
    Impact of step number ($N$) on denoising quality with uniform Schedule.
    Plots of PSNR (left y-axis, red line) and SSIM (right y-axis, blue dashed line) versus the number of steps ($N$) for the proposed 3D Flow model using a uniform sampling schedule ($k=0$) on 1/20 dose of the UDPET dataset. 
    Shaded areas represent standard deviation.
}
\label{fig:metric_step}
\end{figure}

\subsubsection{Comparative Efficiency Benchmarking}
We first benchmarked the inference efficiency and denoising performance (PSNR/SSIM) against representative generative models, including standard 3D DDPM, accelerated 3D DDIM, and DPM-Solver++ applied to the reproduced 3D diffusion baseline.
All evaluations were conducted on the challenging 30 s dose of the FAHZU dataset using identical hardware (single NVIDIA GeForce RTX 4090 GPU).

As detailed in Table~\ref{tab:inference_speed}, our 3D Flow model achieves a notable balance between restoration quality and computational cost. 
Quantitatively, it yields a PSNR of 40.70 $\pm$ 1.68 dB and SSIM of 0.9790 $\pm$ 0.0046, compared with 40.06 $\pm$ 1.66 dB and 0.9741 $\pm$ 0.0060 for 3D DDPM, and 39.72 $\pm$ 1.60 dB and 0.9699 $\pm$ 0.0057 for 200-step 3D DDIM.
This performance is achieved with an inference time of approximately 30 seconds per volume under the standardized inference-only estimate.

The comparison in efficiency reflects both step count and the quality-efficiency trade-off of each sampler.
All methods used the same six-patch sliding-window inference protocol.
The measured 30.74 s runtime for the 2-step 3D Flow model corresponds to 15.37 s per volume-level sampling step, or approximately 2.56 s per patch-level network evaluation including patch processing and volume aggregation overhead.
Under this standardized estimate, the 1000-step 3D DDPM requires approximately 4.27 hours per patient, while 200-step 3D DDIM requires approximately 51.23 minutes.
The 100-step DDIM and 50-step DPM-Solver++ settings are included as aggressive acceleration settings and are not used as the optimized quality baselines because their image quality degraded markedly.
In comparison, our rectified flow model achieves competitive or favorable accuracy using two ODE steps in this benchmark.

\begin{figure*}[!h]
\centerline{\includegraphics[width=1.0\textwidth]{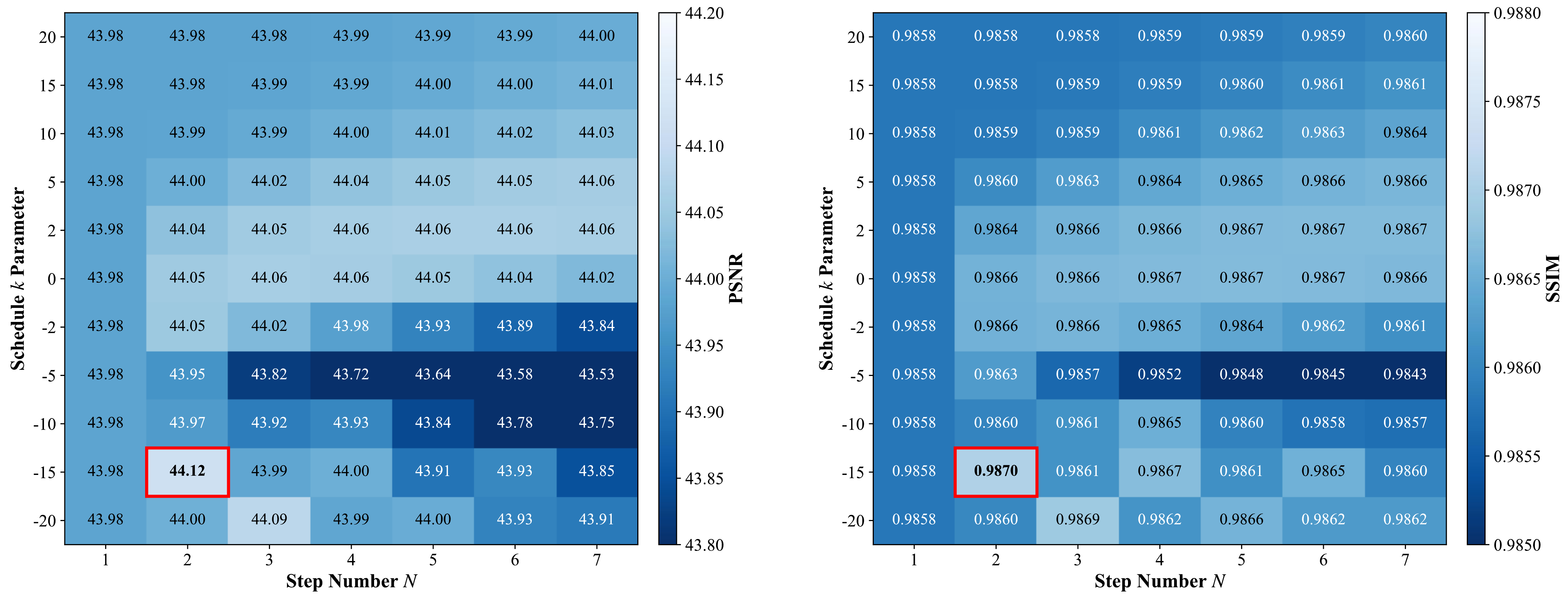}} %
\caption{
    Heatmaps of PSNR and SSIM for varying step number ($N$) and schedule parameter ($k$).
    These heatmaps illustrate the combined impact of the total number of steps ($N$, x-axis) and the non-uniform sampling parameter ($k$, y-axis) on denoising quality (PSNR on left, SSIM on right) for 1/20 dose of the UDPET dataset. 
    Higher values (lighter colors) indicate better performance. 
    The highest performance is highlighted by a red box, achieved at $N$=2 and $k$=-15.
}
\label{fig:heatmap_k_step}
\end{figure*}

\begin{table*}[!htbp]
    \centering
    \caption{\textbf{Ablation of the Non-Uniform Sampling Schedule on UDPET}}
    \label{tab:schedule_ablation}
    \begin{tabular}{lcccccc}
        \toprule
        \textbf{Dose} & \textbf{Uniform PSNR} & \textbf{Proposed PSNR} & \textbf{$\Delta$PSNR} & \textbf{Uniform SSIM} & \textbf{Proposed SSIM} & \textbf{$\Delta$SSIM} \\
        \midrule
        1/100 & 39.39 $\pm$ 2.87 & 40.00 $\pm$ 2.83 & +0.62 & 0.9645 $\pm$ 0.0329 & 0.9702 $\pm$ 0.0285 & +0.0057 \\
        1/50 & 41.64 $\pm$ 2.20 & 42.19 $\pm$ 2.19 & +0.55 & 0.9778 $\pm$ 0.0171 & 0.9811 $\pm$ 0.0148 & +0.0034 \\
        1/20 & 43.83 $\pm$ 2.12 & 44.12 $\pm$ 2.19 & +0.29 & 0.9862 $\pm$ 0.0096 & 0.9871 $\pm$ 0.0090 & +0.0009 \\
        \bottomrule
    \end{tabular}
\end{table*}

Reducing inference from hours to approximately 30 seconds per volume may have practical implications for future deployment. 
The multi-hour processing time of diffusion models typically restricts them to offline, retrospective analysis. 
In contrast, the 30-second inference capability of our 3D Flow model may support efficient online denoising and faster image review while the patient is still on the scanner. 
This capability aligns with the throughput requirements of busy imaging workflows while maintaining favorable image-quality metrics in this study.

\subsubsection{Ablation Study on Sampling Strategy}
The performance of rectified flow is governed by the discretization of the transport trajectory. 
To identify the optimal configuration, we conducted a comprehensive ablation study on the UDPET dataset (1/20 dose), investigating the impact of the step count ($N$) and the time-stepping schedule ($k$).

\subsubsubsection{Impact of Step Count on Uniform Schedule}
To establish a baseline, we evaluated the model's performance (PSNR and SSIM) using a standard uniform time-stepping schedule ($k=0$) while varying the number of steps ($N$) from 1 to 200.
As illustrated in Fig.~\ref{fig:metric_step}, single-step generation ($N=1$) yields suboptimal performance (PSNR $\approx$ 43.98 dB, SSIM $\approx$ 0.9859) despite offering maximum speed. 
This outcome is attributed to significant discretization error, as a single straight linear step provides an insufficient approximation of the curved transport trajectory.
 
Image quality improves as $N$ increases, reaching a peak around $N=3$ (PSNR $\approx$ 44.07 dB, SSIM $\approx$ 0.9867).
Performance remains robust up to approximately $N=7$ steps, beyond which further increases in $N$ led to a gradual decline in image quality. 
This decline is attributed to the accumulation of numerical integration errors over a large number of small steps, which can degrade the overall accuracy of the trajectory approximation despite reducing per-step errors. 
These findings suggest that high reconstruction quality can be achieved within the few-step range.

\subsubsubsection{Joint Optimization of Step and Schedule}
Building upon this finding, we then explored the combined effect of the number of steps ($N$) and our proposed non-uniform sampling parameter ($k$) on denoising quality. 
We evaluated PSNR and SSIM for $N$ values from 1 to 7 and $k$ values ranging from -20 to 20, as presented in the heatmaps in \figref{fig:heatmap_k_step}.

The heatmaps in Fig.~\ref{fig:heatmap_k_step} visualize the sensitivity of PSNR and SSIM to these hyperparameters. 
The optimal configuration was identified at $N=2$ steps with $k=-15$, achieving the highest PSNR (44.12 dB) and SSIM (0.9870).
This finding indicates that a negative $k$ value, such as $k=-15$, which by our function's design concentrates sampling steps towards $t=1$ (denoising target), is beneficial at these very low step counts for PET denoising. 
This strategy may enable the model to allocate computational budget to the final refinement phase, helping suppress residual noise and resolve fine anatomical details where the vector field appears more complex.

To test whether the non-uniform schedule remained beneficial beyond the matched training dose, we further compared the proposed schedule with the uniform schedule at 1/20, 1/50, and 1/100 dose levels on UDPET (Table~\ref{tab:schedule_ablation}).
The comparisons used the same trained model, same cases, same number of integration steps, and same fixed inference initialization, so the schedule was the changed inference variable.
For the uniform-versus-proposed comparisons reported in Table~\ref{tab:schedule_ablation}, paired Wilcoxon signed-rank tests were performed on patient-level PSNR and SSIM values.
All six comparisons, covering three dose levels and two metrics, remained statistically significant after FDR correction ($p<0.001$).
The gains were modest in magnitude but consistent, and were larger under more severe dose reduction.

These results show that selecting both the number of steps and the non-uniform schedule parameter $k$ provides a statistically supported quality-efficiency trade-off for the proposed 3D Flow model.
The optimal evaluated configuration achieved high image-quality metrics (PSNR 44.12 dB, SSIM 0.9870 at 1/20 dose) using two ODE steps.

\begin{figure*}[!tbp] 
    \centering 
    \includegraphics[width=0.95\textwidth]{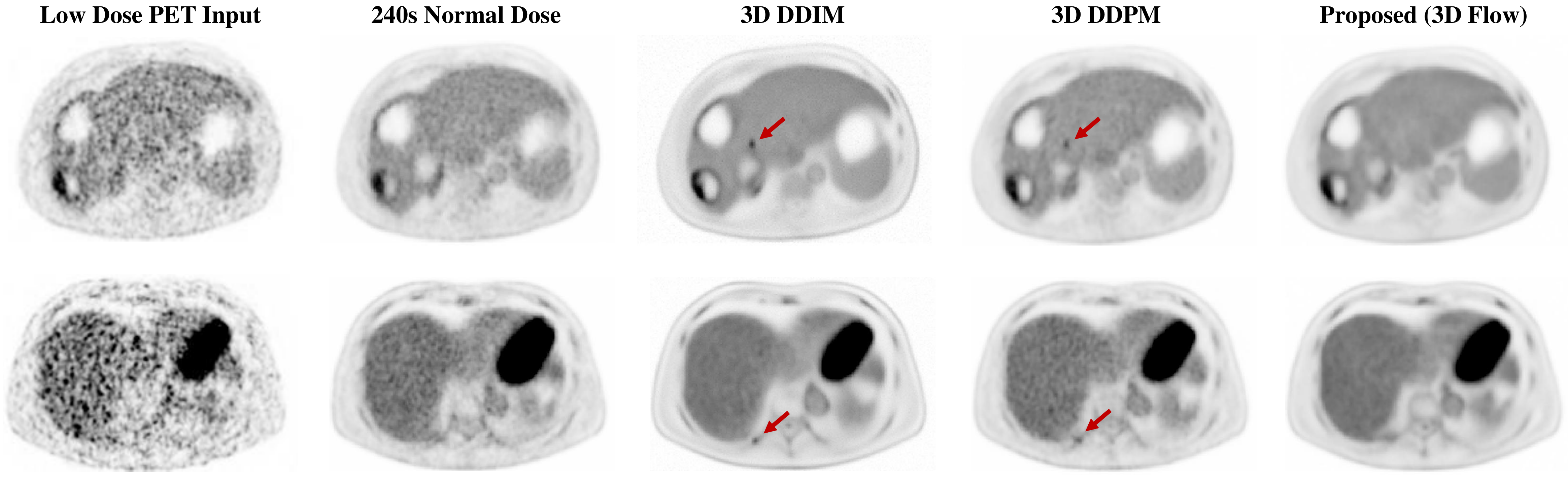} 
    \caption{ 
    Qualitative comparison of focal uptake-like artifacts on axial PET slices from the FAHZU dataset.
    Images display two cases: ultra-low-dose PET input (1st column), 240 s normal-dose reference (2nd), and denoised outputs from 3D DDIM (3rd), 3D DDPM (4th), and proposed 3D Flow model (5th). 
    Red arrows indicate visually apparent focal artifacts or spurious uptake-like patterns in these examples.
    Note that the models mentioned in this figure are trained on UDPET dataset.
    } 
    \label{fig:false_positives}
\end{figure*}

\section{Discussion}
This study introduced and evaluated a 3D rectified flow (3D Flow) model for ultra-low-dose PET image denoising, showing improvements in both image quality metrics and inference efficiency in the tested settings.
Our method demonstrated favorable performance compared with the evaluated generative baselines across challenging low-dose levels and exhibited efficient inference, achieving high image-quality metrics within two integration steps using an empirically optimized non-uniform sampling strategy.
The model showed consistent performance across the evaluated dose levels and data sources without fine-tuning, suggesting potential for cross-condition transfer that should be further validated.

\subsection{Performance Interpretation and Mechanistic Insights}
The observed improvements in PSNR, SSIM, and CNR, particularly at ultra-low-dose levels, suggest improved image quality according to the evaluated metrics.
Specifically, the enhancement in CNR is consistent with improved lesion conspicuity according to this metric, but it should not be interpreted as direct evidence of SUV quantitative accuracy, lesion detectability, or diagnostic equivalence.
Collectively, this performance across varying dose levels suggests that the 3D Flow model learns image transformations that suppress noise while preserving reference-consistent anatomical and functional patterns in the evaluated settings.

The efficacy of our methodology may be related to the intrinsic characteristics of rectified flows and the specific dynamics of the learned trajectory.
The linear-interpolant objective of rectified flow encourages direct transport between noise and target samples, which may be more amenable to low-step numerical integration than long reverse-diffusion chains.
Accordingly, this straightness should be interpreted as an ideal training target, while the learned trajectory can deviate from it in practice.
This interpretation is specific to conditional PET denoising.
SDE-based diffusion models can be advantageous for modeling complex and diverse data distributions, while the present task is a strongly conditional PET denoising problem in which reference consistency, reproducibility, and computational efficiency are central objectives.
To investigate the nuanced dynamics of this transport process, we conducted a high-resolution trajectory analysis using 300 sampling steps. 
This included a dense sampling regime (100 steps) specifically within the final interval $t \in [0.99, 1.0]$ to scrutinize the convergence behavior.

\begin{figure}[!h]
\centering
\includegraphics[width=1.0\columnwidth]{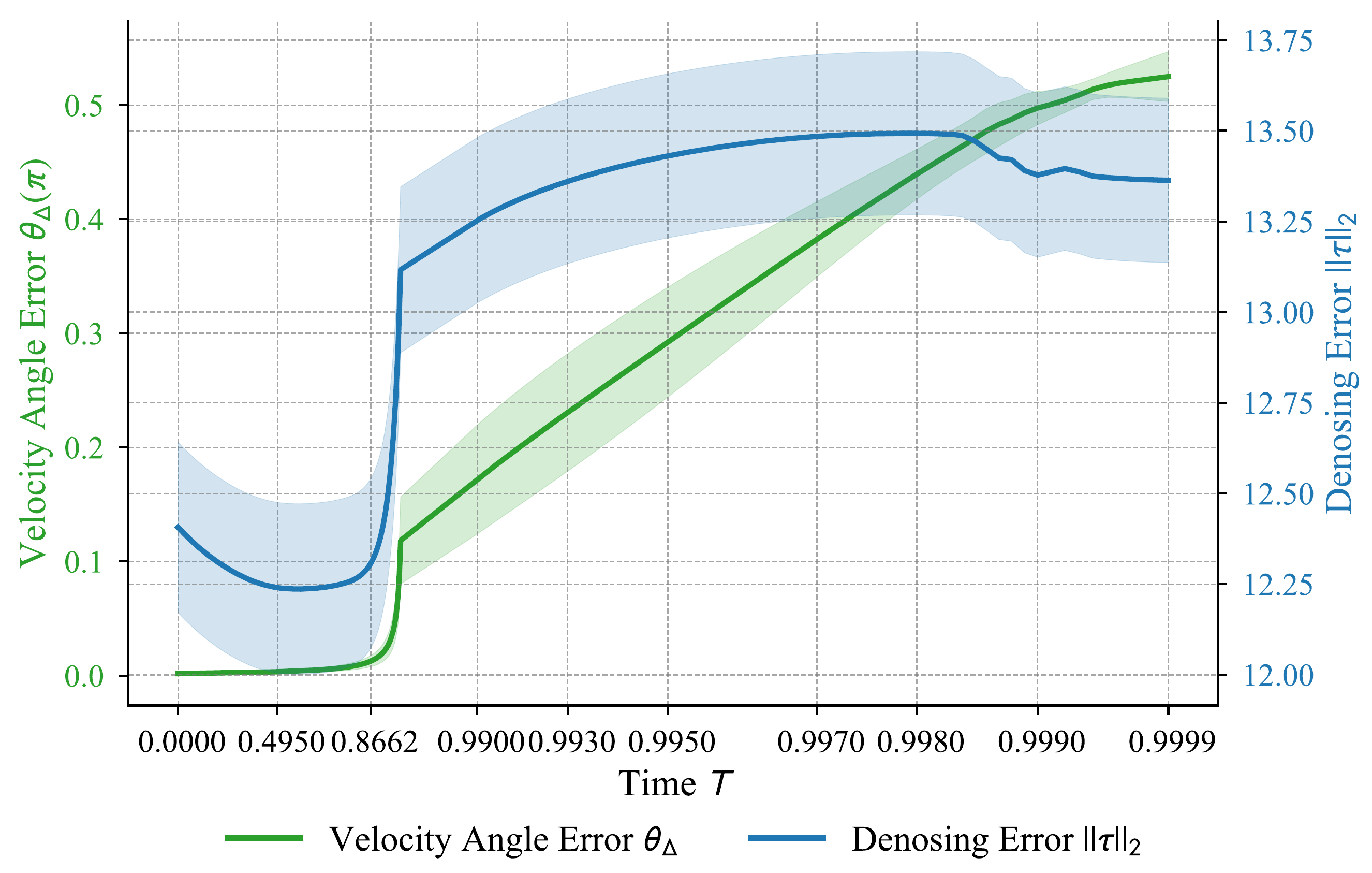}
\caption{
    Plot of velocity angle error ($\theta_\Delta$, green, left y-axis) and denoising error ($||\tau||_2$, blue, right y-axis) over time ($T$, x-axis) for a 3D Flow trajectory. 
    Shaded areas indicate standard deviation. 
    For the denoising error ($\tau$, blue), the shaded area is scaled by a factor of 0.05 for visualization purposes.
}
\label{fig:path_error_analysis}
\end{figure}

We employed two distinct metrics to analyze the learned vector field's characteristics (Fig.~\ref{fig:path_error_analysis}). 
First, the angular error of velocity ($\theta_\Delta$) quantifies the angular deviation between the model's predicted velocity at time $t$ and the ideal velocity vector pointing directly towards the target. 
Second, the denoising error represents the Euclidean distance ($L_2$) between the predicted endpoint and the actual target.

As illustrated in Fig.~\ref{fig:path_error_analysis}, the error profile reveals a notable feature near the trajectory's endpoint. 
For the majority of the path ($t < 0.9$), both errors stay comparatively small, suggesting that the model learns an effective transport direction.
However, as $t$ approaches 1 (denoising target), both the angular error and the denoising error exhibit a sharp increase simultaneously.

This simultaneous surge in both directional (angle) and magnitude (distance) errors indicates larger deviations of the learned vector field from the ideal velocity near the endpoint.
 
Empirically, this suggests that while the model transports the sample to the vicinity of the target, the final convergence may require complex, high-frequency adjustments. 
In this region, deviations in the predicted velocity can lead to overshooting or misdirection if the integration step size is too large.

This observation provides a possible explanation for the motivation of our non-uniform sampling strategy. 
A uniform schedule may apply a relatively coarse step size in this variable $t \approx 1$ region, potentially increasing the integration error. 
Conversely, our optimized schedule (with $k=-15$) concentrates sampling density near the target. 
By reducing the step size as $t \to 1$, the solver can perform finer-grained adjustments while limiting error accumulation.

This alignment between the error profile and the optimal sampling parameter suggests that allocating more integration resolution near the endpoint may help the model achieve favorable reconstruction metrics with few inference steps.
This interpretation is post hoc; the proposed schedule is therefore presented as a fixed globally parameterized schedule selected by validation experiments.

Furthermore, a notable observation was the model's stability in reconstruction. 
As illustrated in Fig.~\ref{fig:false_positives}, outputs from traditional diffusion models like 3D DDPM and 3D DDIM occasionally exhibit spurious uptakes or artifacts that visually resemble focal abnormalities (highlighted by red arrows). 
In contrast, our 3D Flow model's outputs appear smoother and more consistent with the normal-dose reference in these qualitative examples.
This observation should be interpreted as an artifact-focused visual comparison and not as a formal false-positive analysis.
Formal reader studies or lesion-level detection analyses are needed to determine whether these visual differences translate into improved diagnostic interpretation.

\subsection{Clinical Implications and Translational Potential}
The combined attributes of rapid inference and favorable denoising metrics from ultra-low-dose inputs suggest that our 3D Flow model warrants further clinical evaluation. 
The sub-minute processing time, exemplified by 30 seconds per 3D PET volume, could reduce computational delay by enabling faster image reconstruction and review. 
This efficiency may facilitate integration into routine clinical pipelines and could even support more dynamic or efficient online imaging scenarios. 
From a patient perspective, denoising ultra-low-dose acquisitions may contribute to radiation exposure reduction if validated in dedicated clinical studies.
This could contribute to radiation-dose reduction strategies, a particularly important factor for pediatric patients, individuals requiring serial scans, or vulnerable populations. 
Furthermore, shortened scan times have the potential to improve patient comfort and minimize motion artifacts, thereby contributing to higher overall image quality. 
Finally, the model's performance without dataset-specific fine-tuning suggests transfer capabilities that could ease deployment across clinical centers, pending broader validation. 
This may support broader evaluation of advanced PET denoising techniques across different clinical centers.

\subsection{Limitations and Future Works}
Despite its promising results, this study has several limitations that guide future research. 
While our model demonstrated promising performance on the FAHZU and UDPET datasets, its generalizability across an even wider array of PET scanners, diverse radiotracers beyond FDG, and various disease pathologies warrants further investigation. 
Future work should involve extensive multi-center validation on more heterogeneous datasets to fully assess its general applicability.

Additionally, the optimal choice of the non-uniform sampling parameter $k$ was determined empirically through our ablation study. 
Developing a theoretically grounded or adaptive strategy to automatically adjust $k$ and potentially the number of integration steps ($N$) based on the specific characteristics of the input noisy image, or the noise level, would further enhance the model's robustness and ease of use in diverse clinical scenarios. 
Future work should also investigate fast flow-specific solvers and consistency-style generative models after task-specific adaptation to conditional full-volume 3D PET denoising.

Future clinical validation will further benefit from blinded reader studies, formal FROC analysis, lesion-level false-discovery evaluation, SUVmax or SUVmean bias, limits-of-agreement analysis, and reader-assessed diagnostic endpoints.
These analyses would help assess diagnostic equivalence, lesion detectability, absolute uptake preservation, and the clinical impact of visually apparent generative artifacts.
Exploring the integration of our denoising framework with other PET image processing tasks, such as reconstruction, segmentation, or precise quantitative analysis, also presents promising avenues for future research to develop a more comprehensive PET imaging solution.

\section{Conclusion}
This work presents a 3D rectified flow model tailored for ultra-low-dose whole-body PET image denoising. 
The proposed model achieved favorable PSNR, SSIM, and auxiliary CNR-based lesion-conspicuity metrics compared with the evaluated baselines in the tested settings.
A primary contribution is the development of an empirically optimized non-uniform sampling strategy, which calibrates the time schedule parameter ($k$) and uses two integration steps to improve inference efficiency for PET denoising.
This optimization enables our 3D Flow model to achieve favorable reconstruction metrics with approximately 30-second inference per volume in our implementation, compared with multi-hour inference estimates for the evaluated diffusion baseline.
Furthermore, the model exhibits zero-shot transfer performance across the evaluated multi-dose datasets without any fine-tuning, indicating potential adaptability to varying PET imaging protocols within the tested settings.
These findings suggest that our method warrants further validation as a candidate technical approach for ultra-low-dose whole-body PET image denoising.

\bibliographystyle{IEEEtran}

\bibliography{reference}

\begin{thebibliography}{10}
\providecommand{\url}[1]{#1}
\csname url@samestyle\endcsname
\providecommand{\newblock}{\relax}
\providecommand{\bibinfo}[2]{#2}
\providecommand{\BIBentrySTDinterwordspacing}{\spaceskip=0pt\relax}
\providecommand{\BIBentryALTinterwordstretchfactor}{4}
\providecommand{\BIBentryALTinterwordspacing}{\spaceskip=\fontdimen2\font plus
\BIBentryALTinterwordstretchfactor\fontdimen3\font minus \fontdimen4\font\relax}
\providecommand{\BIBforeignlanguage}[2]{{%
\expandafter\ifx\csname l@#1\endcsname\relax
\typeout{** WARNING: IEEEtran.bst: No hyphenation pattern has been}%
\typeout{** loaded for the language `#1'. Using the pattern for}%
\typeout{** the default language instead.}%
\else
\language=\csname l@#1\endcsname
\fi
#2}}
\providecommand{\BIBdecl}{\relax}
\BIBdecl

\bibitem{sweet:1951:uses}
W.~H. Sweet, ``The uses of nuclear disintegration in the diagnosis and treatment of brain tumor,'' \emph{New England Journal of Medicine}, vol. 245, no.~23, pp. 875--878, Dec. 1951.

\bibitem{ming:2020:progress}
Y.~Ming, N.~Wu, T.~Qian, X.~Li, D.~Q. Wan, C.~Li, Y.~Li, Z.~Wu, X.~Wang, J.~Liu, and N.~Wu, ``Progress and future trends in pet/ct and pet/mri molecular imaging approaches for breast cancer,'' \emph{Frontiers in Oncology}, vol.~10, Aug. 2020.

\bibitem{phelps2004molecular}
M.~E. Phelps, ``Molecular imaging and its biological applications,'' \emph{Eur J Nucl Med Mol Imaging}, vol.~31, p. 1544, 2004.

\bibitem{el2011improvement}
G.~El~Fakhri, S.~Surti, C.~M. Trott, J.~Scheuermann, and J.~S. Karp, ``Improvement in lesion detection with whole-body oncologic time-of-flight pet,'' \emph{Journal of Nuclear Medicine}, vol.~52, no.~3, pp. 347--353, 2011.

\bibitem{dutta2013non}
J.~Dutta, R.~M. Leahy, and Q.~Li, ``Non-local means denoising of dynamic pet images,'' \emph{PloS one}, vol.~8, no.~12, p. e81390, 2013.

\bibitem{shidahara2007pet}
M.~Shidahara, Y.~Ikoma, J.~Kershaw, Y.~Kimura, M.~Naganawa, and H.~Watabe, ``Pet kinetic analysis: wavelet denoising of dynamic pet data with application to parametric imaging,'' \emph{Annals of nuclear medicine}, vol.~21, no.~7, pp. 379--386, 2007.

\bibitem{hashimoto2018denoising}
F.~Hashimoto, H.~Ohba, K.~Ote, and H.~Tsukada, ``Denoising of dynamic sinogram by image guided filtering for positron emission tomography,'' \emph{IEEE Transactions on Radiation and Plasma Medical Sciences}, vol.~2, no.~6, pp. 541--548, 2018.

\bibitem{ote2020kinetics}
K.~Ote, F.~Hashimoto, A.~Kakimoto, T.~Isobe, T.~Inubushi, R.~Ota, A.~Tokui, A.~Saito, T.~Moriya, T.~Omura \emph{et~al.}, ``Kinetics-induced block matching and 5-d transform domain filtering for dynamic pet image denoising,'' \emph{IEEE Transactions on Radiation and Plasma Medical Sciences}, vol.~4, no.~6, pp. 720--728, 2020.

\bibitem{suzuki2017overview}
K.~Suzuki, ``Overview of deep learning in medical imaging,'' \emph{Radiological physics and technology}, vol.~10, no.~3, pp. 257--273, 2017.

\bibitem{ronneberger2015u}
O.~Ronneberger, P.~Fischer, and T.~Brox, ``U-net: Convolutional networks for biomedical image segmentation,'' in \emph{International Conference on Medical image computing and computer-assisted intervention}.\hskip 1em plus 0.5em minus 0.4em\relax Springer, 2015, pp. 234--241.

\bibitem{jaudet2021impact}
C.~Jaudet, K.~Weyts, A.~Lechervy, A.~Batalla, S.~Bardet, and A.~Corroyer-Dulmont, ``The impact of artificial intelligence cnn based denoising on fdg pet radiomics,'' \emph{Frontiers in oncology}, vol.~11, p. 692973, 2021.

\bibitem{geng2021content}
M.~Geng, X.~Meng, J.~Yu, L.~Zhu, L.~Jin, Z.~Jiang, B.~Qiu, H.~Li, H.~Kong, J.~Yuan \emph{et~al.}, ``Content-noise complementary learning for medical image denoising,'' \emph{IEEE transactions on medical imaging}, vol.~41, no.~2, pp. 407--419, 2021.

\bibitem{zhu2018image}
B.~Zhu, J.~Z. Liu, S.~F. Cauley, B.~R. Rosen, and M.~S. Rosen, ``Image reconstruction by domain-transform manifold learning,'' \emph{Nature}, vol. 555, no. 7697, pp. 487--492, 2018.

\bibitem{zhou2020supervised}
L.~Zhou, J.~D. Schaefferkoetter, I.~W. Tham, G.~Huang, and J.~Yan, ``Supervised learning with cyclegan for low-dose fdg pet image denoising,'' \emph{Medical image analysis}, vol.~65, p. 101770, 2020.

\bibitem{fu2023aigan}
Y.~Fu, S.~Dong, M.~Niu, L.~Xue, H.~Guo, Y.~Huang, Y.~Xu, T.~Yu, K.~Shi, Q.~Yang \emph{et~al.}, ``Aigan: Attention--encoding integrated generative adversarial network for the reconstruction of low-dose ct and low-dose pet images,'' \emph{Medical Image Analysis}, vol.~86, p. 102787, 2023.

\bibitem{ho2020denoising}
J.~Ho, A.~Jain, and P.~Abbeel, ``Denoising diffusion probabilistic models,'' \emph{Advances in neural information processing systems}, vol.~33, pp. 6840--6851, 2020.

\bibitem{yu2025robust}
B.~Yu, S.~Ozdemir, Y.~Dong, W.~Shao, T.~Pan, K.~Shi, and K.~Gong, ``Robust whole-body pet image denoising using 3d diffusion models: evaluation across various scanners, tracers, and dose levels,'' \emph{European Journal of Nuclear Medicine and Molecular Imaging}, vol.~52, no.~7, pp. 2549--2562, 2025.

\bibitem{song2020denoising}
J.~Song, C.~Meng, and S.~Ermon, ``Denoising diffusion implicit models,'' \emph{arXiv preprint arXiv:2010.02502}, 2020.

\bibitem{liu2022rectified}
Q.~Liu, ``Rectified flow: A marginal preserving approach to optimal transport,'' \emph{arXiv preprint arXiv:2209.14577}, 2022.

\bibitem{zhu:2024:flowie}
Y.~Zhu, W.~Zhao, A.~Li, Y.~Tang, J.~Zhou, and J.~Lu, ``Flowie: Efficient image enhancement via rectified flow,'' in \emph{2024 IEEE/CVF Conference on Computer Vision and Pattern Recognition (CVPR)}, Jun. 2024, pp. 13--22.

\bibitem{qin:2025:reversing}
H.~Qin, W.~Luo, L.~Wang, D.~Zheng, J.~Chen, M.~Yang, B.~Li, and W.~Hu, ``Reversing flow for image restoration,'' Jun. 2025.

\bibitem{ohayon:2025:posteriormean}
G.~Ohayon, T.~Michaeli, and M.~Elad, ``Posterior-mean rectified flow: Towards minimum mse photo-realistic image restoration,'' Feb. 2025.

\bibitem{song2023consistencymodels}
\BIBentryALTinterwordspacing
Y.~Song, P.~Dhariwal, M.~Chen, and I.~Sutskever, ``Consistency models,'' 2023. [Online]. Available: \url{https://arxiv.org/abs/2303.01469}
\BIBentrySTDinterwordspacing

\bibitem{salimans2022progressivedistillationfastsampling}
\BIBentryALTinterwordspacing
T.~Salimans and J.~Ho, ``Progressive distillation for fast sampling of diffusion models,'' 2022. [Online]. Available: \url{https://arxiv.org/abs/2202.00512}
\BIBentrySTDinterwordspacing

\bibitem{kim2024simplereflowimprovedtechniques}
\BIBentryALTinterwordspacing
B.~Kim, Y.-G. Hsieh, M.~Klein, M.~Cuturi, J.~C. Ye, B.~Kawar, and J.~Thornton, ``Simple reflow: Improved techniques for fast flow models,'' 2024. [Online]. Available: \url{https://arxiv.org/abs/2410.07815}
\BIBentrySTDinterwordspacing

\bibitem{sun2025pet}
Y.~Sun and O.~Mawlawi, ``Pet image denoising with rectified flow,'' 2025.

\bibitem{ma2025newoneshotfederatedlearning}
\BIBentryALTinterwordspacing
Y.~Ma, H.~Zhang, Q.~Yang, G.~Luo, and Y.~Zhu, ``A new one-shot federated learning framework for medical imaging classification with feature-guided rectified flow and knowledge distillation,'' 2025. [Online]. Available: \url{https://arxiv.org/abs/2507.19045}
\BIBentrySTDinterwordspacing

\bibitem{yazdani2025flowmatchingmedicalimage}
\BIBentryALTinterwordspacing
M.~Yazdani, Y.~Medghalchi, P.~Ashrafian, I.~Hacihaliloglu, and D.~Shahriari, ``Flow matching for medical image synthesis: Bridging the gap between speed and quality,'' 2025. [Online]. Available: \url{https://arxiv.org/abs/2503.00266}
\BIBentrySTDinterwordspacing

\bibitem{lipman2022flow}
Y.~Lipman, R.~T. Chen, H.~Ben-Hamu, M.~Nickel, and M.~Le, ``Flow matching for generative modeling,'' \emph{arXiv preprint arXiv:2210.02747}, 2022.

\bibitem{Kuangyu2022Ultra}
\BIBentryALTinterwordspacing
{Kuangyu Shi}, R.~Guo, S.~Xue, A.~Rominger, and B.~Li, ``Ultra-low {Dose} {PET} {Imaging} {Challenge} 2022,'' mar 16 2022. [Online]. Available: \url{https://zenodo.org/record/6361846}
\BIBentrySTDinterwordspacing

\bibitem{alberts2021clinical}
I.~Alberts, J.-N. H{\"u}nermund, G.~Prenosil, C.~Mingels, K.~P. Bohn, M.~Viscione, H.~Sari, B.~Vollnberg, K.~Shi, A.~Afshar-Oromieh \emph{et~al.}, ``Clinical performance of long axial field of view pet/ct: a head-to-head intra-individual comparison of the biograph vision quadra with the biograph vision pet/ct,'' \emph{European journal of nuclear medicine and molecular imaging}, vol.~48, no.~8, pp. 2395--2404, 2021.

\bibitem{wang2004image}
Z.~Wang, A.~C. Bovik, H.~R. Sheikh, and E.~P. Simoncelli, ``Image quality assessment: from error visibility to structural similarity,'' \emph{IEEE transactions on image processing}, vol.~13, no.~4, pp. 600--612, 2004.

\bibitem{cui2019pet}
J.~Cui, K.~Gong, N.~Guo, C.~Wu, X.~Meng, K.~Kim, K.~Zheng, Z.~Wu, L.~Fu, B.~Xu \emph{et~al.}, ``Pet image denoising using unsupervised deep learning,'' \emph{European journal of nuclear medicine and molecular imaging}, vol.~46, no.~13, pp. 2780--2789, 2019.

\bibitem{Wilcoxon1992}
\BIBentryALTinterwordspacing
F.~Wilcoxon, \emph{Individual Comparisons by Ranking Methods}.\hskip 1em plus 0.5em minus 0.4em\relax New York, NY: Springer New York, 1992, pp. 196--202. [Online]. Available: \url{https://doi.org/10.1007/978-1-4612-4380-9_16}
\BIBentrySTDinterwordspacing

\bibitem{xie2023ddpet3ddoseawarediffusionmodel}
\BIBentryALTinterwordspacing
H.~Xie, W.~Gan, B.~Zhou, X.~Chen, Q.~Liu, X.~Guo, L.~Guo, H.~An, U.~S. Kamilov, G.~Wang, and C.~Liu, ``Ddpet-3d: Dose-aware diffusion model for 3d ultra low-dose pet imaging,'' 2023. [Online]. Available: \url{https://arxiv.org/abs/2311.04248}
\BIBentrySTDinterwordspacing

\bibitem{jiang2025fast}
H.~Jiang, M.~Imran, T.~Zhang, Y.~Zhou, M.~Liang, K.~Gong, and W.~Shao, ``Fast-ddpm: Fast denoising diffusion probabilistic models for medical image-to-image generation,'' \emph{IEEE Journal of Biomedical and Health Informatics}, 2025.

\end{thebibliography}

\end{document}